%% file: main.tex
\documentclass{article}
\usepackage{arxiv} 

\usepackage{graphicx}%
\usepackage{amsmath,amssymb,amsfonts}%
\usepackage{amsthm}%
\usepackage{xcolor}%
\usepackage{manyfoot}%

\usepackage{booktabs}%
\usepackage{xspace}
\usepackage{stmaryrd}

\usepackage{tikz}
\usetikzlibrary{shapes.geometric,arrows,positioning,fit,patterns,shapes,decorations.pathreplacing,calc}
\usepackage{forest}

\usepackage{url}
\usepackage[hidelinks]{hyperref}
\usepackage{cleveref}
\definecolor{ourcolsGray}{gray}{0.6}
\definecolor{ourcolsLightGray}{gray}{0.9}
\definecolor{ourcolsLineGray}{gray}{0.3}

\newcommand{\sr}{\textsc{Savile Row}\xspace}
\newcommand{\af}{\textsc{AutoFolio}\xspace}
\newcommand*{\email}[1]{\href{mailto:#1}{\nolinkurl{#1}} } 

\renewcommand{\shorttitle}{Learning to Select SAT Encodings}
\renewcommand{\headeright}{Accepted Manuscript}
\renewcommand{\undertitle}{Accepted Manuscript}

\begin{document}

\title{Learning to Select SAT Encodings for Pseudo-Boolean and Linear Integer
  Constraints}

\date{Accepted for Publication September 2023}

\author{
  Felix Ulrich-Oltean\\
  \email{felix.ulrich-oltean@york.ac.uk}
  \And
  Peter Nightingale\\
  \email{peter.nightingale@york.ac.uk}
  \And
  James Alfred Walker\\
  \email{james.walker@york.ac.uk}
  \And
  Department of Computer Science, University of York\\
  Deramore Lane, York, YO10 5GH, United Kingdom
  }

\maketitle

\begin{abstract}Many constraint satisfaction and optimisation problems can be solved
  effectively by encoding them as instances of the Boolean Satisfiability
  problem (SAT).  However, even the simplest types of constraints have many
  encodings in the literature with widely varying performance, and the problem
  of selecting suitable encodings for a given problem instance is not trivial.
  We explore the problem of selecting encodings for pseudo-Boolean and linear
  constraints using a supervised machine learning approach. We show that it is
  possible to select encodings effectively using a standard set of features for
  constraint problems; however we obtain better performance with a new set of
  features specifically designed for the pseudo-Boolean and linear
  constraints. In fact, we achieve good results when selecting encodings for
  unseen problem classes.  Our results compare favourably to AutoFolio when
  using the same feature set.  We discuss the relative importance of instance
  features to the task of selecting the best encodings, and compare several
  variations of the machine learning method.
\end{abstract}


\section{Introduction}

Many constraint satisfaction and optimisation problems can be solved effectively
by encoding them as instances of the Boolean Satisfiability problem
(SAT). Modern SAT solvers are remarkably effective even with large formulas, and
have proven to be competitive with (and often faster than) CP solvers (including
those with conflict learning).  However, even the simplest types of constraints
have many encodings in the literature with widely varying performance, and the
problem of predicting suitable encodings is not trivial.

We explore the problem of selecting encodings for constraints of the form
\(\sum_{i=1}^n q_ie_i\diamond k\) where \(\diamond \in \{<,\leq,=,\neq,\geq,>\}\),
\(q_1\ldots q_n\) are integer coefficients, \(k\) is an integer constant and
\(e_i\) are decision variables or simple expressions containing a single decision
variable (such as negation). 
We separate these constraints into two classes:
\emph{pseudo-Boolean} (PB) when all \(e_i\) are Boolean; and
\emph{linear integer} (LI) when there exists an integer expression \(e_i\).  We
treat these two classes separately, selecting one encoding for each class when
encoding an instance. 

We select from a set of state-of-the-art encodings, including all eight
encodings of Bofill et
al~\cite{bofillCompactMDD2017,bofill2019sat,bofillPBAMO2022} which are
extensions of the Generalized Totalizer~\cite{joshiGenTot2015}, Binary Decision
Diagram~\cite{abioNewLookBDDs2012}, Global Polynomial
Watchdog~\cite{bailleuxPB2009}, Local Polynomial Watchdog~\cite{bailleuxPB2009},
Sequential Weight Counter~\cite{holldoblerCompactPB2012}, and n-Level Modulo
Totalizer~\cite{zha2019n}. All eight of these encodings are for pseudo-Boolean
constraints combined with at-most-one (AMO) sets of terms (where at most one of
the corresponding \(e_i\) Boolean expressions are true in a solution). The AMO sets come from
an integer variable or are detected automatically~\cite{ansocp2019autodetect} as
described in \Cref{sec:satenc}.  
We also use an encoding named \emph{Tree} which is described in this paper. 

The context for this work is \sr~\cite{nightingaleAutoCon2017}, a constraint
modelling tool that takes the modelling language Essence Prime and can produce
output for various types of solver, including CP, SAT, and recently
SMT~\cite{davidsonEffectiveSMT2020}. When encoding a constraint to SAT, two
different approaches may be taken depending on the type of constraint.  Some
constraint types are decomposed into simpler constraints prior to encoding
(e.g.\ allDifferent is decomposed into a set of at-most-one constraints, stating
that each relevant domain value appears at most once).  Other constraint types
are encoded to SAT directly, in which case \sr will apply the encoding chosen on
the command-line (or the default if no choice is made).

We use a supervised machine learning approach, trained with a corpus of 614
instances from 49 problem classes (constraint models).  We show that it is
possible to select encodings effectively, approaching the performance of the
virtual best encoding (i.e.\ the best possible choice for each instance), using
an existing set of features for constraint problem instances. Also we obtain
better performance by adding a new set of features specifically designed for the
pseudo-Boolean and linear integer constraints, especially when selecting
encodings for unseen problem classes.

We study two versions of the encoding selection problem: \textit{split-by-instance} and 
\textit{split-by-class}. In the first, the set of all 
instances is split into training and test sets with no reference to the 
problem classes (in common with earlier work \cite{hurleyProteus2014,stojadinovicMeSAT2014}). 
For any given test instance, the model may have been trained on other instances of the same class. 
In \textit{split-by-class}, each test instance belongs to a problem class that was
not seen in training. We would argue that \textit{split-by-class} is a realistic
and useful version of the encoding selection problem because it is desirable for
constraint modelling tools to be robust for new problem classes. 
Even when the encoding and solver settings will be hand-optimised for an important new problem class, a
good initial configuration is likely to be useful. 

\subsection{Contributions}

In summary, our contributions are as follows:
\begin{itemize}
    \item We address the problem of selecting SAT encodings for instances of
      \emph{unseen} problem classes, which we argue is a realistic version of
      the encoding selection problem.  To our knowledge, all previous approaches
      (such as \cite{hurleyProteus2014,stojadinovicMeSAT2014}) train and test
      their machine learning models on instances drawn from the same set of
      problem classes.
    \item We describe a machine learning approach that produces very good
      results, and that performs much better than the mature, self-tuning
      algorithm selection tool \af~\cite{lindauerAutoFolio2015}.
    \item We present a new set of features for pseudo-Boolean and linear integer
      constraints, and show improved overall performance and robustness when
      using them.
    \item We evaluate our machine learning method thoroughly, and present an
      analysis of feature importance.
    \item We describe \sr's \emph{Tree} encoding in detail, expanding on
      the summary given in \cite{ulrichSelectingCP2022}.
\end{itemize}
 
  \paragraph{Note}
  This paper extends the earlier conference paper \cite{ulrichSelectingCP2022} in several
  ways.  First, the set of encodings has been extended from 5 to 9, and now includes all 8
  from a very recent work on encoding pseudo-Boolean constraints \cite{bofillPBAMO2022}.
  Secondly, we give a more precise and complete background regarding the SAT encodings,
  including a full description of the \emph{Tree} encoding that is only summarised
  elsewhere.  Finally, we have substantially extended the analysis and discussion of
  experimental results.

\subsection{Preliminaries}\label{sec:prelim}

A \emph{constraint satisfaction problem} (CSP) is defined as a set of variables
$X$, a function that maps each variable to its domain,
$D : X \rightarrow 2^{\mathbb{Z}}$ where each domain is a finite set, and a set
of constraints $C$. A \emph{constraint} $c \in C$ is a relation over a subset of
the variables $X$. The {\em scope} of a constraint $c$, named
$\textrm{scope}(c)$, is the set of variables that $c$ constrains. A
\emph{constraint optimisation problem} (COP) also minimises or maximises the
value of one variable. A \emph{solution} is an assignment to all variables that
satisfies all constraints $c \in C$.

Boolean
Satisfiability (SAT) is a subset of CSP with only Boolean variables and only
constraints (\emph{clauses}) of the form \((l_1\vee \cdots \vee l_k)\) where
each \(l_i\) is a literal \(x_j\) or \(\neg x_j\).  A \emph{SAT encoding} of a
CSP variable \(x\) is a set of SAT variables and set of clauses with exactly one
solution for each value in \(D(x)\). A SAT encoding of a constraint \(c\) is a
set of clauses and additional Boolean variables \(A\), where the clauses contain
only literals of \(A\) and of the encodings of variables in
$\textrm{scope}(c)$. An encoding of \(c\) has \emph{at least} one solution
corresponding to each solution of \(c\).  Also, an encoding of \(c\) has \textit{no} 
solutions corresponding to a non-solution of \(c\). Literals of \(A\) may only appear 
within the encoding of \(c\).  A more sophisticated definition of constraint encoding
would allow variables in \(A\) to be shared among multiple constraint encodings,
however none of the encodings used in this paper share additional variables. 

\emph{Generalised arc consistency}
(GAC) for a constraint $c$ means that for a given partial assignment, all values
are removed from the domain of each variable in $\textrm{scope}(c)$ if they
cannot appear in any extended assignment satisfying $c$.  A SAT encoding of
\(c\) has the property \emph{GAC} iff unit propagation of the SAT
encoding of \(c\) results in the following correspondence: for each variable
\(x_i \in \textrm{scope}(c)\), the set of remaining solutions of the encoding of \(x_i\)
corresponds to the set of values in \(D(x_i)\) after GAC has been enforced on
\(c\).

A SAT encoding of \(c\) has the \textit{Consistency Checker} (CC)
property~\cite{bofillPBAMO2022} iff unit propagation of the SAT encoding of
\(c\) will derive \textit{false} when the SAT partial assignment corresponds to
a CSP partial assignment that cannot be extended to a full assignment that
satisfies \(c\).


\section{Learning to Choose SAT Encodings}

First we describe the palette of encodings for PB and LI constraints, then our
approach to selecting encodings using instance features and machine learning.

\subsection{SAT Encodings}\label{sec:satenc}

Recall that we are considering constraints of the following form where
$q_1\ldots q_n$ are integer coefficients, $k$ is an integer constant and $e_i$
are decision variables or simple expressions containing one variable.
\begin{equation*}
\sum_{i=1}^n q_ie_i\diamond k \quad \mathrm{where}\quad \diamond \in \{<, \leq, =, \neq, \geq, >\}
\end{equation*}
An expression \(e_i\) may be: an integer or Boolean variable \(x_i\); a negated
Boolean variable \(\neg x_i\); or a comparison \((x_i \# k_i)\) where
\(\#\in \{<,\leq,=,\neq, \geq, >\}\), \(x_i\) is an integer variable or Boolean
literal, and \(k_i\) is a constant.  We refer to the set of values that \(e_i\)
can take as \(D(e_i)\), extending the notation \(D(x_i)\).  We distinguish
between \textit{top-level} constraints that must be satisfied in all solutions,
and \textit{nested} constraints that are contained in a logic operator such as
\(\vee\) or \(\rightarrow\).

Initial normalisation steps are applied to all constraints of this form,
regardless of the choice of encoding. All \(<\) constraints are converted to
\(\leq\), and \(>\) constraints to \(\geq\) by adjusting \(k\).  If the
coefficients \(q\) have a greatest common divisor (GCD) greater than 1 then all
coefficients are divided by the GCD. When the comparator is \(\leq\) then
\(k\leftarrow \lfloor k/\mathrm{GCD} \rfloor \), and when the comparator is
\(\geq\) then \(k\leftarrow \lceil k/\mathrm{GCD} \rceil \). When
\(\diamond \in \{=,\neq \}\), \(k\leftarrow k/\mathrm{GCD}\) unless
\(k/\mathrm{GCD}\) is non-integer in which case the constraint is evaluated to
\textit{true} or \textit{false} as appropriate.

If the comparator \(\diamond\) is \(\neq\) then a new auxiliary variable \(a\)
is created whose domain is the set of all values the sum may take. A new
top-level LI equality constraint is introduced, as follows.
\begin{equation*}
\sum_{i=1}^n q_ie_i -a = 0
\end{equation*}
The original constraint is replaced with \(a\neq k\), and if it is top-level
then \(k\) will be removed from the domain of \(a\).

Any constraints that are not top-level (i.e.\ are nested in another expression
such as a disjunction) are always encoded with the \emph{Tree} encoding in \sr,
described in \Cref{sub:treeenc}.  For top-level constraints, we have nine
encodings available and each can be applied to either PB or LI constraints,
producing 81 configurations.

\subsubsection{SAT Encoding Preliminaries}

First we describe the encoding of CSP variables in \sr. Boolean variables and
integer variables that have two values are encoded with a single SAT
variable. For other integer variables, \sr\ generates the
\emph{direct}~\cite{dekleer1989comparison} or
\emph{order}~\cite{crawford-baker-jobshop} encoding (or both) as required to
encode the constraints on that variable. We provide these details for
completeness and without claiming novelty.

The direct encoding of a variable \(x\) has one SAT variable representing each
value in \(D(x)\).  It provides a SAT literal, or a constant \textit{true} or
\textit{false}, for a proposition \((x=a)\) or \((x\neq a)\) for any integer
\(a\), with the literal denoted \(\llbracket x=a\rrbracket\) or
\(\llbracket x\neq a\rrbracket \).  We use the 2-product encoding
\cite{chen2010new} to ensure \(x\) is assigned at most one (AMO) value, and a
single clause to ensure \(x\) is assigned at least one (ALO) value.

The order encoding of \(x\) introduces one SAT variable for each of the
following propositions.
\begin{equation*}
(x\leq v)\quad \forall v \in (D(x)\setminus \{\mathrm{max}(D(x))\})
\end{equation*}
The order encoding provides a SAT literal (or constant \textit{true} or
\textit{false}) for a proposition \((x\leq a)\) or its negation \((x>a)\) for
any integer \(a\), with the literal denoted \(\llbracket x\leq a\rrbracket\) or
\(\llbracket x>a \rrbracket\).  For each pair of values
\(\{v_1, v_2\}\subseteq (D(x)\setminus \{\mathrm{max}(D(x))\})\) where
\(v_1<v_2\) and \(\nexists v'\in D(x): v_1<v'<v_2\), a clause is generated to
ensure coherence of the encoding:
\begin{equation*}
\llbracket x\leq v_1\rrbracket \rightarrow \llbracket x\leq v_2\rrbracket
\end{equation*}
When the direct and order encodings are both required they are channelled
together with the following set of clauses. The AMO constraint of the direct
encoding is omitted in this case.
\begin{gather*}
  \llbracket x\leq v \rrbracket \wedge  \llbracket x > v-1 \rrbracket \rightarrow \llbracket x=v \rrbracket, \\
  \llbracket x=v \rrbracket \rightarrow \llbracket x\leq v \rrbracket, \quad
  \llbracket x=v \rrbracket \rightarrow \llbracket x > v-1 \rrbracket \quad
  \forall v \in D(x)
\end{gather*}
There are two equivalences that can be used to slightly improve the direct-order
encoding:
\(\llbracket x\leq a \rrbracket \leftrightarrow \llbracket x=a \rrbracket\) for
minimum value \(a\), and
\(\llbracket x> b-1 \rrbracket \leftrightarrow \llbracket x=b \rrbracket\) for
maximum value \(b\). By using these two equivalences we can remove two SAT
variables so \(2\lvert D(x)\rvert -3\) variables are generated.

\subsubsection{PB(AMO) Encodings}

\sr\ implements nine SAT encodings for linear and pseudo-Boolean
constraints. Eight of these are encodings of PB(AMO)
constraints~\cite{bofillCompactMDD2017,bofill2019sat,bofillPBAMO2022}, which are
pseudo-Boolean constraints with non-intersecting at-most-one (AMO) groups of
terms (where at most one of the corresponding \(e_i\) expressions are true in
any solution). Encodings of PB(AMO) constraints can be substantially smaller and
more efficient to solve than the corresponding PB constraints
\cite{bofillCompactMDD2017,bofill2019sat,ansocp2019autodetect,bofillPBAMO2022}.

For the eight PB(AMO) encodings the constraints must be placed in a normal form
where all coefficients are positive, only \(\leq\) is allowed, and each \(e_i\)
must be a Boolean expression.  The first normalisation step is to decompose
equality into two inequalities \(\leq\) and \(\geq\).  Second, any \(\geq\)
constraints are converted to \(\leq\) by negating all \(q_i\) coefficients and
\(k\).  At this stage, all constraints are in \(\leq\) form.

Non-trivial integer terms \(q_ie_i\) (where \(e_i\) is not Boolean and
\( D(e_i) \neq \{0,1\}\)) are dealt with as follows. If \(q_i<0\) then
\(q_i \leftarrow -q_i\) and \(e_i \leftarrow -e_i\).  The integer variable
contained in \(e_i\) is required to have a \textit{direct} SAT encoding.
Finally, \(q_ie_i\) is replaced with an AMO group of \(\lvert D(e_i)\rvert-1\)
terms representing each value of \(q_ie_i\) except the smallest (which is
cancelled out by adjusting \(k\)). For example, if \(q_i=2\) and \(e_i\) has
values \(\{1,2,3\}\) then \(k\leftarrow k-2\) and \(q_ie_i\) would be replaced
with the following AMO group of two terms.
\begin{equation*}
2(e_i=2)+4(e_i=3)
\end{equation*}

At this point all terms \(q_ie_i\) in the constraint must be Boolean or have
domain \(D(e_i) = \{0,1\}\). If \(q_i<0\), the term is replaced with
\(-q_i(e_i=0)\) or \(-q_i(\neg e_i)\) (as appropriate for the type of \(e_i\))
and \(k\leftarrow k-q_i\). Otherwise, the term is replaced with \(q_i(e_i=1)\)
or remains unchanged (as appropriate for the type).

Automatic AMO detection~\cite{ansocp2019autodetect} (which applies constraint
propagation to find AMO groups among the Boolean terms of the original
constraint) is enabled in our experiments. Automatic AMO detection has been
shown to substantially improve solving time in some
cases~\cite{ansocp2019autodetect}. It partitions the Boolean terms of the
constraint into AMO groups, and (as described above) each integer term becomes
an AMO group also.  Finally, we exactly follow the PB(AMO) normalisation rules
of Bofill et al~\cite{bofillPBAMO2022} prior to encoding.

The PB(AMO) encodings are as follows:

\begin{description}
\item[MDD]The Multi-valued Decision Diagram
  encoding~\cite{bofillCompactMDD2017} (a generalisation of the BDD encoding for
  PB constraints~\cite{abioNewLookBDDs2012}) uses an MDD to encode the PB(AMO)
  constraint. Each layer of the MDD corresponds to one AMO group. BDDs and MDDs
  are a popular choice for encoding sums to SAT since they can compress
  equivalent states in each layer.

\item[GGPW] The Generalized Global Polynomial Watchdog
  encoding~\cite{bofill2019sat} (generalising GPW~\cite{bailleuxPB2009}) 
  is based on bit arithmetic and is polynomial in size. It has the CC property but not GAC. 
  
\item[GLPW] The Generalized Local Polynomial Watchdog
  encoding~\cite{bofillPBAMO2022} (generalising LPW~\cite{bailleuxPB2009}) is
  similar to GGPW but has the GAC property.  However, while being polynomial in
  size, it is often too large to be practical.

\item[GGT] The Generalized Generalized Totalizer~\cite{bofill2019sat}
  encodes the PB(AMO) constraint with a binary tree, where the leaves represent
  the AMO groups and each internal node represents the sum of all leaves beneath
  it. GGT extends the Generalized Totalizer~\cite{joshiGenTot2015}. The binary
  tree is constructed using the minRatio heuristic~\cite{bofillPBAMO2022} that
  aims to minimise the numbers of values of internal nodes.

\item[GGTd] is identical to GGT except that a balanced binary tree is used. 
  
\item[RGGT] The Reduced Generalized Generalized Totalizer~\cite{bofillPBAMO2022} 
attempts to improve on GGT by compressing equivalent states at its internal nodes. 
The minRatio heuristic is used to construct the tree. 

\item[GSWC] The Generalized Sequential Weight Counter~\cite{bofill2019sat}
  (based on the Sequential Weight Counter~\cite{holldoblerCompactPB2012})
  encodes the sum of each prefix sub-sequence of the AMO groups.

\item[GMTO] The Generalized n-Level Modulo
  Totalizer~\cite{bofillPBAMO2022} (based on the n-Level Modulo
  Totalizer~\cite{zha2019n}) is an extension of the Generalized Totalizer that
  represents values of the internal nodes in a mixed-radix base. GMTO can be
  substantially more compact than other PB(AMO) encodings.
\end{description}

The MDD, GLPW, GGT, GGTd, RGGT, and GSWC encodings all have the GAC property
(described in \cref{sec:prelim}). Where the original constraint \(c\) contains
no integer terms and the comparator of the original constraint is neither \(=\)
nor \(\neq\) then these six encodings will enforce GAC on \(c\).  GGPW has the
CC property. Where the original constraint \(c\) contains no integer terms and
the comparator of the original constraint is not \(=\) or \(\neq\) then GGPW
will have the CC property for \(c\).

\subsubsection{Tree Encoding}\label{sub:treeenc}

The \emph{Tree} encoding is related to the Generalized Totalizer (GT)
\cite{joshiGenTot2015} encoding of PB constraints. However, Tree is able to
natively encode equalities (in addition to inequalities), and it natively
supports integer variables and negative coefficients.  Tree does not support
arbitrary AMO groups of terms (i.e.\ it is not a PB(AMO) encoding) so it does
not benefit from automatic AMO detection~\cite{ansocp2019autodetect}.

Given a constraint \(c: \sum_{i=1}^n q_ie_i\diamond k\) where
$\diamond \in \{\leq, =, \geq\}$, the first step is to normalise each term to
have a lower bound of 0. Each term \(q_ie_i\) is shifted such that its smallest
value becomes 0, and \(k\) is adjusted accordingly.  Shifting a term \(q_ie_i\)
is implemented with a linear view that introduces no SAT variables or clauses
when \(c\) is encoded.

The encoding process has two main stages. In the first stage, a tree is
constructed from \(c\), with each term (integer or Boolean) represented by a
leaf.  The tree is binary with the exception of the root node which may have
three children. Each internal non-root node has a corresponding auxiliary
integer variable that represents the sum of its two child nodes. A constraint is
generated to connect each internal non-root node to its two children.  The
process to construct the tree maintains a set \(S\) of terms, initially
containing all terms in the original sum. While \(S\) contains more than three
terms, two terms \(w_iy_i\) and \(w_jy_j\) are removed from \(S\) and replaced
with one new term \(a\), where \(a\) is a new auxiliary variable. A constraint
\(w_iy_i+w_jy_j-a\diamond 0\) is generated.  Finally \(S\) contains at most
three terms and the last constraint is generated: \(\sum S \diamond k\).

For each new variable \(a\) (introduced for the sum of \(w_iy_i\) and
\(w_jy_j\)), the domain of \(a\) is the set of values obtainable by adding any
value of \(w_iy_i\) to any value of \(w_jy_j\). If \(\diamond \in \{\leq, =\}\),
values greater than \(k\) are removed from the domain of \(a\).  The shape of
the tree and domain sizes of the internal variables are determined by the choice
of terms to remove from \(S\) at each step. Tree uses a simple heuristic that
chooses a term with the smallest number of possible values, breaking ties in
favour of the smallest range from minimum to maximum value. The aim of the
heuristic is to minimise the number of values of the auxiliary variables.  To
complete the first stage, if \(\diamond\) is \(=\) then each equality constraint
is broken down into one \(\leq\) and one \(\geq\) constraint.

As an example, consider the following constraint (where each variable  \(x_i\) is Boolean):
\begin{equation}
20x_1+30x_2+20x_3+40x_4+10x_5+20x_6+x_7 \leq 55\label{eq-treecons}
\end{equation}

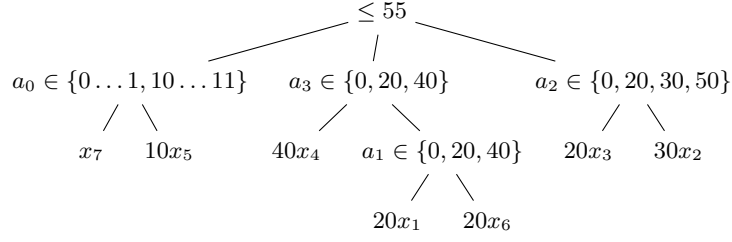
\begin{figure}
\centering
\footnotesize
\begin{forest}
for tree = {s sep=9pt, align=center}
[{$ \le 55$}
  [{$a_0 \in \{0 \ldots 1,10 \ldots 11\}$}
    [{$x_7$}]
    [{$10 x_5$}]
  ]
  [{$a_3 \in \{0,20,40 \}$}
    [{$40 x_4$}]
    [{$a_1 \in \{ 0,20,40 \}$}
      [{$20 x_1$}]
      [{$20 x_6$}]
    ]
  ]
  [{$a_2 \in \{0,20,30,50 \}$}
    [{$20 x_3$}]
    [{$30 x_2$}]
  ]
]
\end{forest}
\caption{The tree generated for the \textit{Tree} encoding of $20x_1+30x_2+20x_3+40x_4+10x_5+20x_6+x_7 \leq 55$}
\label{fig:tree-enc}
\end{figure}

The tree generated for this constraint is shown in \Cref{fig:tree-enc}. In this
case the terms all have two possible values and the heuristic generates a
balanced tree. Four auxiliary variables are generated as shown in
\Cref{fig:tree-enc}, and five constraints are generated as follows.
\begin{equation*}
\begin{aligned}
x_7+10x_5-a_0 & \leq 0\\
20x_1+20x_6-a_1 & \leq 0\\
20x_3+30x_2-a_2 & \leq 0\\
40x_4+a_1-a_3 & \leq 0\\
a_0+a_2+a_3 & \leq 55
\end{aligned}
\end{equation*}
The second stage is to encode the variables and constraints to SAT.  The order
encoding is required for integer leaf nodes and is used for all \(a_i\)
variables on the internal nodes.  Constraints are encoded using an improved
version of the original order encoding of Tamura et
al~\cite{tamura2009compiling}. The original order encoding builds a set of
clauses from the intervals
\(\{ \mathrm{min}(D(q_ie_i))-1\ldots\mathrm{max}(D(q_ie_i))\}\) (for all
\(i\)). Tamura, Banbara, and Soh \cite{tamura2013compiling} presented an
improved order encoding that uses the sets \(D(q_ie_i)\) and is more compact
than the original version.  The version presented here also uses the sets
\(D(q_ie_i)\), and by taking the entire set into account it avoids generating
redundant clauses compared to the original version.  First, terms are sorted by
increasing number of possible values. For a constraint of arity \(r\), the set
of clauses is as follows:
\begin{equation*}
\bigwedge_{\langle b_1\ldots b_r\rangle \in B} \bigvee_i \left\{ \begin{aligned}
& \llbracket e_i \leq (b_i/q_i)-1 \rrbracket \quad & \mathrm{when}\quad (q_i>0 \wedge i<r) \\
& \llbracket e_i > (b_i/q_i) \rrbracket \quad & \mathrm{when}\quad (q_i<0 \wedge i<r) \\
& \llbracket e_i \leq \lfloor b_i/q_i \rfloor \rrbracket \quad & \mathrm{when}\quad (q_i>0 \wedge i=r) \\
& \llbracket e_i > \lceil b_i/q_i \rceil -1 \rrbracket \quad & \mathrm{when}\quad (q_i<0 \wedge i=r) 
\end{aligned}
\right\}
\end{equation*}

where the set of tuples \(B\) is defined as follows.

\begin{equation*}
B=\left\{ \langle b_1 \ldots b_r \rangle \mid \forall_{i=1}^{r-1} b_i\in D(q_ie_i) \quad \wedge \quad b_r=k-\sum_{i=1}^{r-1} b_i  \right\}
\end{equation*}

For example, we encode the first constraint as follows (where false literals and
one trivially true clause have been removed).
\begin{gather*}
\llbracket x_7\leq 0 \rrbracket \vee \llbracket a_0> 0\rrbracket \\
\llbracket x_5\leq 0\rrbracket \vee \llbracket a_0> 1\rrbracket \\
\llbracket x_7\leq 0\rrbracket \vee \llbracket x_5\leq 0\rrbracket \vee \llbracket a_0> 10\rrbracket
\end{gather*}
The original order encoding~\cite{tamura2009compiling} would generate two
additional (redundant) clauses, as follows.
\begin{gather*}
\llbracket x_5\leq 0\rrbracket \vee \llbracket a_0> 0\rrbracket \\
\llbracket x_7\leq 0\rrbracket \vee \llbracket x_5\leq 0\rrbracket \vee \llbracket a_0> 1\rrbracket
\end{gather*}

When applied to a PB \(\leq\) constraint, Tree is similar to GT in most
respects. The main difference occurs at the root node, which for Tree has 3
children and no auxiliary SAT variables, while for GT the root node has 2
children and an auxiliary SAT variable for each possible value of the sum that
lies between 0 and \(k+1\).  On the example constraint, Tree generates 10 SAT
variables (in addition to the original \(x_1\ldots x_7\)) and 30 clauses. GT
with the minRatio heuristic generates 20 SAT variables and 36 clauses, while GT
with a balanced binary tree produces 21 SAT variables and 54 clauses.

When applied to an equality constraint, Tree generates one set of auxiliary
variables at each internal (non-root) node and these are used to encode both
\(\leq\) and \(\geq\) parts of the constraint. In contrast, when using any
PB(AMO) encoding, the constraint is decomposed into two inequalities which are
then encoded entirely separately. As a result Tree can be more compact in terms
of SAT variables. Replacing \(\leq\) with \(=\) in \Cref{eq-treecons}, the Tree
encoding introduces 10 variables and 53 clauses, whereas GT with minRatio
generates 48 variables and 82 clauses.

Tree has the GAC property when the comparator of the original constraint is
neither \(=\) nor \(\neq\).

\subsubsection{Discussion}

The set of 9 encodings is diverse but not exhaustive. Ab\'io et al proposed a
BDD-based encoding for linear constraints~\cite{abio2015encoding}, however it
has been directly related to the MDD encoding~\cite{bofillMDDPBAMO2020}.  In
addition to MDD-based encodings, Ab\'io et al propose two further encodings for
linear constraints~\cite{abioEncLinCons2020}: one based on sorting networks
(SN), which is related to the GPW encoding, and another log-based encoding
BDD-Dec.  Other log encodings such as the one used by
Picat-SAT~\cite{zhou2017optimizing} may also be more effective in some cases.

For our experiments we use an extended version of \sr
1.9.1~\cite{srManual190}. All constraints other than PB and LI use the default
encoding as described in the \sr manual.

\subsection{Instance Features}

Our task of selecting the best SAT encodings relies on extracting features of
constraint problems in order to predict a performant encoding configuration. We
initially use existing generic CSP instance features, and then go on to define
features which relate directly to the PB and LI constraints in a given
CSP instance.  The resulting featuresets are described below.

\label{sec:features}
\begin{description}
\item[f2f] We use the \texttt{fzn2feat} tool~\cite{amadiniFeatEx2014} to
  extract 95 static instance features relating to the number and types of
  variables and their domains, the types and sizes of constraints and features
  of the objective in optimisation problems. The full list of features can be
  found at \url{https://github.com/CP-Unibo/mzn2feat}; some features were not
  applicable, e.g. there are no float variables in Essence Prime and \sr does
  not produce all the same annotations.  The \texttt{fzn2feat} tool requires
  \texttt{FlatZinc} models as input -- we generate these using \sr's
  standard \texttt{FlatZinc} back-end.
\item[f2fsr] We also re-implement the \emph{f2f} features as closely
  as possible within \sr, applied to the model directly before encoding to SAT.
\item[lipb] We introduce a new set of 45 features describing the PB
  constraints in a problem instance. We also extract these for LI constraints,
  giving 90 new features in total. These features are listed in
  \Cref{tab:features}.
\item[combi] We combine the \emph{f2fsr} and \emph{lipb} features.
\end{description}

\begin{table}[!hbt]
\caption{New features for pseudo-Boolean and linear integer constraints. For
  each aspect of a constraint listed in the left column, we calculate the
  aggregates in the right column.  In the aggregation functions, \emph{IQR}
  means inter-quartile range, \emph{skew} refers to the non-parametric skew and
  \emph{ent} is Shannon's entropy.  The identifier for each aspect is given in
  brackets.}
\label{tab:features}
\centering
\setlength{\tabcolsep}{3pt}
\footnotesize
\renewcommand{\arraystretch}{1}
\begin{tabular}{@{}p{0.6\textwidth}p{0.4\textwidth}@{}}
\toprule
Aspect of constraint & Aggregate\\
\midrule
Number of (PB or LI) constraints (\texttt{count}) & none \\
Number of terms (\texttt{n}) & min, max, mean, median, IQR, skew, ent, sum \\
Sum of coefficients (\texttt{wsum}) & sum, skew, IQR \\
Minimum coefficient (\texttt{q0}) & min, mean \\
Maximum coefficient (\texttt{q4}) & max, median, mean \\
Median coefficient (\texttt{q2}) & median, skew, ent \\
IQR of coefficients (\texttt{iqr}) & median, skew \\
Coefficients' quartile skew (\texttt{skew}) & mean, min, max, ent \\
Distinct coefficient values (\texttt{sep}) & mean, max \\
Ratio of distinct coeff. values to num. of coeffs. (\texttt{sepr}) & mean, max \\
Number of At-Most-One groups (AMOGs) (\texttt{amogs}) & mean \\
Mean size of AMO group (\texttt{asize\_mn}) & mean \\
Mean AMOG size $ \div $  number of terms (\texttt{asize\_r2n}) & mean  \\
Mean maximum coeff. size in AMOGs (\texttt{amaxw\_mn}) & mean  \\
Skew of maximum coeff. in AMOGs (\texttt{amaxw\_skew}) & mean, ent \\
Upper limit ($k$) (\texttt{k}) & mean, median, max, IQR, ent, skew\\
$k$ $\times$ number of AMOGs (\texttt{k\_amo\_prod}) & mean, IQR, ent \\
\bottomrule
\end{tabular}
\end{table}

The rationale behind the \emph{lipb} features listed in \Cref{tab:features} is
to represent the distribution and partitioning of coefficients within LI and PB
constraints.  For example, by inspecting averages, quartiles and measures of
skew, we can distinguish between PBs with terms mostly similar in weight and
ones where one or two coefficients dominate.  Additionally, because most
encodings we use make use of AMO groups, our features also consider the
characteristics of the AMO groups in a constraint.  Of course we lose the
individual detail when we aggregate over all constraints in the instance, but by
using a variety of aggregation functions we hope to represent both the
characteristics of the individual constraints and how they are distributed
within the instance as a whole.

\subsection{Problem Corpus}

We begin with the 65 constraint models with a total of 757 instances from a
recent paper~\cite{davidsonEffectiveSMT2020} in order to work with a wide
variety of problem classes.  An added advantage is that the models are written
in Essence Prime, \sr's input language.  Unfortunately this collection has a
very skewed distribution of instances per problem class, ranging from just 1 to
100.  We mitigate this in two ways: firstly, we limit the number of instances
per class to 50 by taking a random sample where more instances are available;
secondly, we add instances to existing classes where it is easy, such as when
instance parameters are just a few integers.  We also add two problem classes
from recent XCSP3 competitions~\cite{2019XCSP3Competition2019}: the Balanced
Academic Curriculum Problem (BACP) and the Hamiltonian Cycle Problem (HCP).
Some of the problems are filtered out during the data cleaning phase; we give
details of this process and the resulting corpus in \Cref{sec:datasets} and
\Cref{tab:instances}.  

\subsection{Training}
\label{sec:training}
We evaluated several classifier models from the \texttt{scikit-learn}
library~\cite{scikit-learn}, including decision trees and forests of extremely
randomised trees, as well as the XGBoost classifier~\cite{chenXGBoost2016}.  We
also investigated training various regressors to predict runtime.  We found that a
random forest classifier performs best for our purposes.  The
\texttt{scikit-learn} implementation is based on Breiman's random
forests~\cite{breimanRandomForests2001}, but uses an average of predicted
probabilities from its decision trees rather than a simple vote.

We follow the method of Probst et al.~\cite{probstRFHPTuning2019} who
investigated hyperparameter tuning for random forests and concluded that the
number of estimators should be set sufficiently high (we use 200) and that it is
worth tuning the \emph{number of features}, \emph{maximum tree depth}, and
\emph{sample size}.  We use randomised search with 50 iterations and 5-fold
cross-validation to tune the hyperparameters.  We experimented with more tuning
iterations but it did not lead to improved prediction quality.

If a classifier makes a poor prediction, the consequences vary.  It is possible
that the chosen encodings lead to a running time which is very close to that of
the ideal choice; the opposite is also true and misclassification can be very
expensive.  To address this issue, we follow a similar approach to the
\emph{pairwise classification} used in \af~\cite{lindauerAutoFolio2015}: we
train a random forest model for each of the possible pairs of encoding
configurations.  When making predictions, each model chooses between its two
candidates.  The configuration with most votes is chosen; if two or more
configurations have equal votes, we select the one which produced the shortest
total running time over the training set.  This approach effectively creates a
predicted ranking of configurations from the features and often leads to better
prediction performance than using a single random forest classifier.

\begin{figure}[tb]
\centering
\includegraphics[width=0.9\textwidth]{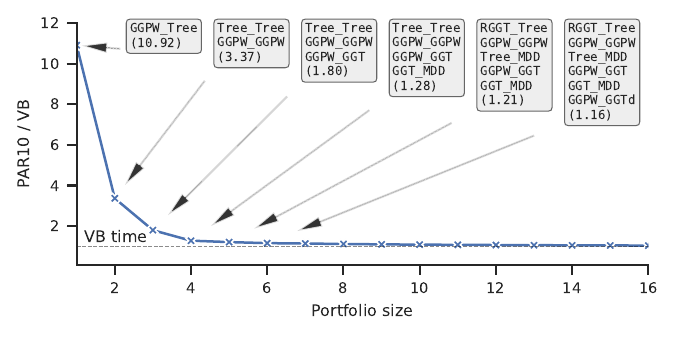}
\caption{The virtual best PAR10 run-time on our whole corpus for a range of
  portfolio sizes, as a multiple of the overall virtual best; the resulting
  portfolios (of \emph{li\_pb} configurations) are shown for sizes 1 to 6.}
\label{fig:portfolio-times}
\end{figure}

To facilitate the pairwise training and prediction approach, we reduce our
selection of encoding combinations from 81 (9 PB encodings $\times$ 9 LI
encodings) to a portfolio of 6, thus needing to train just 15 models (rather
than 3240 if we had used all 81 choices).  We seek to retain performance
complementarity as described in \cite{kerschkeAlgoSel2019} from a much reduced
portfolio size.  The portfolio is built from the timings in the training set
using a greedy approach as follows. We begin with a single encoding
configuration in the portfolio and then successively add the remaining
configuration which would lower the virtual best PAR10 time by the biggest
margin (PAR10 is defined in \Cref{sec:datasets}).  We do this until we have a
portfolio of 6.  We repeat the process using each of the 81 configurations as
the starting element and finally select the best-performing portfolio from these
81.  \Cref{fig:portfolio-times} shows that this portfolio reduction has a small
impact on the virtual best performance across our corpus -- the virtual best
time for a portfolio of size 6 is within 16\% of the time achievable with all
configurations.

In addition to the pairwise voting scheme, we implement two further
customisations when training the classifiers:
\begin{description}
\item[Sample Weights] Firstly we aim to give more importance to
  instances which are harder (with a longer virtual best runtime) and where the
  encoding choice makes a bigger difference.  Each instance is given a positive
  integer weight $w$ according to the formula
  \[
    w=\lfloor \log_{10}{(10 + t_{VB} \times \frac{t_{VW}}{t_{VB}})} \rfloor
  \]
  where $t_{VB}$ and $t_{VW}$ are the very best and very worst runtimes
  respectively for the instance.
\item[Custom Loss] Secondly, we provide a custom loss function for
  the cross-validation used during hyperparameter tuning.  The custom loss
  function simply returns the difference in time between the runtime of the
  chosen encoding configuration and the virtual best for that instance.

\end{description}

To conduct a more complete comparison we also implement two additional
alternative setups.
\begin{description}
\item[Single Classifier] We try using a single random forest classifier
  with the same portfolio of 6 configurations (combining PB and LI encodings).
  In terms of hyperparameter tuning, we try it both with a generous allowance of
  15 times the iterations allocated to each classifier in the pairwise setup,
  and with a more restricted budget of 60 iterations. 
\item[Separate LI/PB Choice] Secondly we modify the pairwise setup to
  make a separate prediction for LI and PB constraints, choosing a portfolio of
  6 encodings for each encoding type.  This approach has its difficulties
  because when labelling the dataset with the best encoding for one type of
  constraint, the encoding of the other constraint type must be chosen somehow.
  We address this by setting the other constraint type to the single best for
  the training set. This setup is more expensive in terms of training time,
  effectively repeating the entire process for each constraint type under
  consideration, rather than taking advantage of a complementary portfolio
  across two (or more) encodings.
\end{description}


\section{Empirical Investigation}

\subsection{Method}
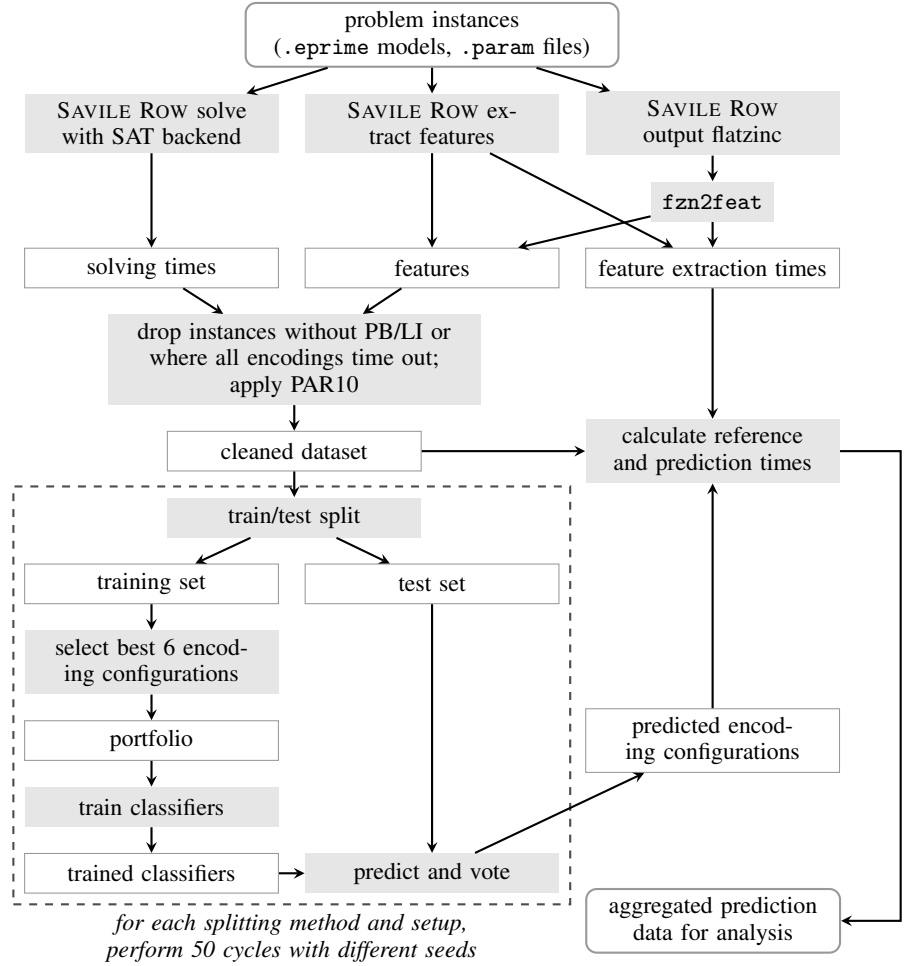
\begin{figure}[tb]
  \centering
  \footnotesize
  \input{diagram}
  \caption{An overview of the steps involved in our experimental investigation.
  The white boxes with solid borders represent data; the grey boxes represent
  processes.}
  \label{fig:overview}
\end{figure}

We provide an overview of our experimental process in~\Cref{fig:overview}.
Briefly, the method consists of:
\begin{itemize}
  \item Running \sr with different encoding choices in order to collect runtime
    information and to extract features.
  \item Cleaning the resulting dataset.
  \item Carrying out 50 \emph{split, train, predict} cycles with each of our
    machine learning setups, using the same train/test splits in order to allow
    fair comparison across the setups.
  \item Using the predicted encoding choices to identify the resulting runtimes.
  \item Aggregating the ``predicted'' runtimes and calculating reference times
    for comparison.
\end{itemize}

The experimental design is described in more detail below.

\subsubsection{Solving Problem Instances and Extracting Features}

We run \sr\ on each instance in the corpus with each of the 81 encoding
configurations.  The CNF clause limit is set to 5 million and the \sr\ time-out
to 1 hour.  We switch on automatic detection of At-Most-One
constraints~\cite{ansocp2019autodetect}.  We choose Kissat as our SAT solver as
it formed the basis of the top three performers in the 2021 SAT
competition~\cite{SATCompetitions}.  We use the latest release available at the
time, \texttt{sc2021-sweep}~\cite{proceedingsSATComp2021}, with default settings
and separate time limit of 1 hour.  The experiment is run on the Viking research
cluster with Intel Xeon 6138 20-core 2.0 GHz processors; we set the memory limit
for each job to 8 GB.  We carry out 5 runs (with distinct random seeds) for each
configuration to average out stochastic behaviour of the solver.

To extract the features we run each problem instance once with the \sr\ feature
extractor and once to generate standard FlatZinc (using the \texttt{-flatzinc}
flag) followed by \texttt{fzn2feat}~\cite{amadiniFeatEx2014}.  We record the
time taken to extract the features.

\subsubsection{Cleaning the Dataset}\label{sec:datasets}

We calculate the median runtime over 5 runs for each instance and encoding
configuration.  We mark a result as timed out if the total runtime (\sr $+$
Kissat) exceeds 1~hour. To decide what penalty to apply to runs which time out,
we consider all instances for whom every configuration finishes within the
allocated time.  The mean \emph{worst/best} ratio is 13.06 and the median ratio
is 4.91.  When we consider only those problem instances which are not solved
with any encoding configuration in less than 1 minute, then the
\emph{worst/best} mean ratio is 9.18 so we believe it fair to penalise a
time-out by a factor of 10.  We therefore choose to use PAR10, i.e.\ assigning
10~hours to any result which takes longer than our 1 hour time-out limit.  This
is the same penalty applied in other related literature~\cite{hurleyProteus2014,
  lindauerAutoFolio2015, stojadinovicMeSAT2014} which addresses the problem of
selecting SAT encodings for CSPs.

Having applied PAR10, we filter the corpus as follows.  We drop instances if
they contain no PB or LI constraints.  We also exclude any instances which end
up requiring no SAT solving -- \sr\ can sometimes solve a problem in
pre-processing through its automatic re-formulation and domain filtering.
Finally we exclude instances for which all configurations time out.  At this
point, 614 instances of 49 problem classes remain in the corpus;
\Cref{tab:instances} shows the number of instances for each problem class and
the mean number of PB and LI constraints per instance.

\begin{table}[!t]
\caption{Number of instances \emph{(\#)}, mean number of PB constraints
  \emph{(PBs)} and mean number of LI constraints \emph{(LIs)} per
  instance for each problem class in the eventual corpus.}
\label{tab:instances}
\centering
\setlength{\tabcolsep}{6pt}
\footnotesize
\begin{tabular}{@{}lrrrlrrr@{}}
  \toprule
  Problem Class & \# & PBs & LIs & Problem Class & \# & PBs & LIs\\
  \midrule

killerSudoku2 & 50 & 1811.2 & 129.9 & carSequencing & 49 & 435.7 & 0.0\\
knights & 44 & 170.5 & 336.9 & langford & 39 & 146.2 & 0.0\\
opd & 36 & 21.9 & 76.2 & knapsack & 28 & 1.0 & 1.0\\
sonet2 & 24 & 10.0 & 1.0 & immigration & 23 & 0.0 & 1.0\\
bibd-implied & 22 & 410.6 & 0.0 & efpa & 21 & 162.8 & 0.0\\
handball7 & 20 & 705.0 & 1206.0 & mrcpsp-pb & 20 & 90.0 & 45.7\\
n\_queens & 20 & 1593.0 & 0.0 & bibd & 19 & 338.7 & 0.0\\
briansBrain & 16 & 0.0 & 1.0 & life & 16 & 0.0 & 438.9\\
molnars & 16 & 0.0 & 4.0 & n\_queens2 & 16 & 309.0 & 0.0\\
bpmp & 14 & 14.0 & 0.0 & blackHole & 11 & 202.2 & 0.0\\
pegSolitaireTable & 8 & 59.9 & 0.0 & pegSolitaireState & 8 & 59.9 & 0.0\\
pegSolitaireAction & 8 & 59.9 & 0.0 & magicSquare & 7 & 136.0 & 36.0\\
peaceArmyQueens1 & 7 & 0.0 & 1008.0 & peaceArmyQueens3 & 6 & 0.0 & 4.0\\
quasiGrp5Idem & 6 & 586.7 & 0.0 & golomb & 6 & 59.2 & 38.7\\
quasiGrp7 & 6 & 410.7 & 0.0 & quasiGrp6 & 6 & 410.7 & 0.0\\
quasiGrp4NonIdem & 4 & 1067.5 & 208.0 & quasiGrp3NonIdem & 4 & 1067.5 & 208.0\\
quasiGrp5NonIdem & 4 & 389.0 & 0.0 & quasiGrp4Idem & 4 & 416.0 & 208.0\\
bacp & 4 & 0.0 & 25.0 & quasiGrp3Idem & 4 & 458.0 & 208.0\\
waterBucket & 4 & 0.0 & 46.0 & discreteTomography & 2 & 240.0 & 0.0\\
solitaire\_battleship & 2 & 72.0 & 16.0 & plotting & 1 & 1.0 & 28.0\\
nurse & 1 & 27.0 & 42.0 & grocery & 1 & 0.0 & 2.0\\
farm\_puzzle1 & 1 & 0.0 & 2.0 & diet & 1 & 0.0 & 6.0\\
sokoban & 1 & 0.0 & 24.0 & sonet & 1 & 3.0 & 1.0\\
contrived & 1 & 0.0 & 4.0 & sportsScheduling & 1 & 166.0 & 64.0\\
tickTackToe & 1 & 6.0 & 14.0 &  &  &  & \\
  \bottomrule
\end{tabular}
\end{table}

\subsubsection{Splitting the Corpus, Training and Predicting}
\label{sec:splits}

For each of our classifier setups and our four featuresets, we run a
\emph{split, train, predict} cycle 50 times.  We use seeds 1 to 50 to
co-ordinate the splits so that we compare the prediction power of the different
feature sets and setups using the same training and test sets.

For each cycle, we aim for an 80:20 train to test split using two approaches.
The \emph{split-by-instance} approach simply selects instances at random with
uniform probability -- with this approach, instances of any given problem class
are usually found in both the training and test sets.  The \emph{split-by-class}
approach also splits problems randomly but ensures that all instances of a
problem class end up either in the training or the test set, meaning that
predictions are being made on unseen problem classes.  This second method can
lead to the test set being slightly larger than 20\%.

Prior to training the classifiers, the portfolio of available configurations is
built based on the runtimes of the training set.  Then the training instances are
labelled for each pairwise classifier with the configuration that has the
shorter runtime.  For each pairwise classifier, we search the hyperparameter
space and fit the model to the training set.  Finally, we make predictions using
the test set ready for evaluation.

\subsubsection{Evaluating the Performance of Predicted Encodings}
\label{sec:evaluating}

To evaluate the impact of using the learnt encoding choices, we calculate two
benchmarks commonly used in algorithm selection~\cite{kerschkeAlgoSel2019}: the
\emph{Virtual Best (VB)} time is the total time taken to solve the instances in
a test set if we always made the best possible choice from all 81
configurations; and the \emph{Single Best (SB)} time is the total
time taken to solve the instances in a test set when using the single
configuration that minimises the total solving time on the training set.  In
addition we refer to: the time taken using {\sr}'s \emph{default (Def)}
configuration, which is the \emph{Tree} encoding for both PB and LI constraints,
and finally the \emph{Virtual Worst (VW)} time to indicate the overall variation
in performance of the encoding configurations in the portfolio.

\begin{table}[thp]
\caption{Total PAR10 times over the 50 test sets as a multiple of the virtual
  best configuration time.  The best time for each combination of setup,
  features and splitting method is shown in \textbf{bold}.  The predicted
  runtimes include feature extraction time.  In the setup details, \emph{Co/Sep}
  shows whether LI and PB encodings were selected separately or as a combined
  choice; \emph{SW} means sample weighting is used; \emph{CL} indicates custom
  loss used in cross-validation; \emph{Tuning} refers to the number of cycles of
  hyperparameter tuning, or the time budget in the case of \af.}
\label{tab:timings}
\centering
\setlength{\tabcolsep}{6pt}

\begin{tabular}{@{}rrrr@{}}
  \multicolumn{4}{c}{\emph{Split-by-Instance Reference Times}}\\
  \toprule
  Virtual Best & Single Best & Default & Virtual Worst \\
  \midrule
  1.00 & 13.87 & 17.84 & 123.70\\
  \bottomrule
  \addlinespace[1em]
\end{tabular}

\begin{tabular}{@{}lcccrrrrr@{}}
  \multicolumn{9}{c}{\emph{Split-by-Instance Predicted Times}}\\
  \toprule
  \multicolumn{5}{c}{Setup Details}&\multicolumn{4}{c}{Relative Times by Featureset}\\
  \cmidrule(r){1-5}\cmidrule(l){6-9}
  Selector & Co/Sep & SW & CL & Tuning & f2f & f2fsr & lipb & combi \\
  \cmidrule(r){1-5}\cmidrule(l){6-9}
  Pairwise Voting & co & - & - & $50 \times 15$ & 7.99 & \textbf{5.63} & 5.87 & 5.86 \\
  Pairwise Voting & co & \checkmark & - & $50 \times 15$ & 6.33 & 5.36 & 5.10 & \textbf{4.99} \\
  Pairwise Voting & co & - & \checkmark & $50 \times 15$ & 6.01 & 4.74 & \textbf{4.57} & 4.75 \\
  Pairwise Voting & co & \checkmark & \checkmark & $50 \times 15$ & 5.40 & 4.69 & \textbf{4.43} & 4.57 \\
  \cmidrule(r){1-5}\cmidrule(l){6-9}
  Single Classifier & co & \checkmark & \checkmark & 750 & 3.91 & \textbf{3.70} & 3.77 & 3.81 \\
  Single Classifier & co & \checkmark & \checkmark & 60 & 3.95 & \textbf{3.70} & 3.98 & 3.83 \\
  Single Classifier & co & - & - & 750  & 10.34 & 9.64 & 9.11 & \textbf{8.90} \\
  \cmidrule(r){1-5}\cmidrule(l){6-9}
  Pairwise Voting & sep & \checkmark & \checkmark & $50 \times 15 \times 2$ & 6.53 & 6.67 & \textbf{5.60} & 5.81 \\
  \cmidrule(r){1-5}\cmidrule(l){6-9}
  \af & co & n/a & n/a & 1 hour & 22.19 & 24.18 & \textbf{20.63} & 21.02 \\
  \af & co & n/a & n/a & 2 hours & 26.71 & 27.59 & 22.63 & \textbf{22.19} \\
  \af & co & n/a & n/a & 4 hours & 22.76 & 25.40 & \textbf{22.45} & 23.47 \\
  \bottomrule
  \addlinespace[1em]
\end{tabular}

\begin{tabular}{@{}rrrr@{}}
  \multicolumn{4}{c}{\emph{Split-by-Class Reference Times}}\\
  \toprule
  Virtual Best & Single Best & Default & Virtual Worst \\
  \midrule
  1.00 & 25.40 & 17.15 & 160.96\\
  \bottomrule
  \addlinespace[1em]
\end{tabular}

\begin{tabular}{@{}lcccrrrrr@{}}
  \multicolumn{9}{c}{\emph{Split-by-Class Predicted Times}}\\
  \toprule
  \multicolumn{5}{c}{Setup Details}&\multicolumn{4}{c}{Relative Times by Featureset}\\
  \cmidrule(r){1-5}\cmidrule(l){6-9}
  Selector & Co/Sep & SW & CL & Tuning & f2f & f2fsr & lipb & combi \\
  \cmidrule(r){1-5}\cmidrule(l){6-9}
  Pairwise Voting & co & - & - & $50 \times 15$ & 15.07 & 15.01 & 14.17 & \textbf{12.55} \\
  Pairwise Voting & co & \checkmark & - & $50 \times 15$ & 16.80 & 14.90 & 13.21 & \textbf{12.77} \\
  Pairwise Voting & co & - & \checkmark & $50 \times 15$ & 16.42 & 15.30 & 14.97 & \textbf{11.16} \\
  Pairwise Voting & co & \checkmark & \checkmark & $50 \times 15$ & 15.69 & 15.19 & \textbf{11.00} & 11.84 \\
  \cmidrule(r){1-5}\cmidrule(l){6-9}
  Single Classifier & co & \checkmark & \checkmark & 750 & 20.59 & 19.15 & \textbf{13.17} & 14.53 \\
  Single Classifier & co & \checkmark & \checkmark & 60 & 22.01 & 19.49 & 13.67 & \textbf{13.52} \\
  Single Classifier & co & - & - & 750 & 19.72 & 19.93 & \textbf{15.18} & 16.50 \\
  \cmidrule(r){1-5}\cmidrule(l){6-9}
  Pairwise Voting & sep & \checkmark & \checkmark & $50 \times 15 \times 2$ & 16.91 & 14.10 & \textbf{12.02} & 12.71 \\
  \cmidrule(r){1-5}\cmidrule(l){6-9}
  \af & co & n/a & n/a & 1 hour & 24.28 & 27.31 & \textbf{24.21} & 26.79 \\
  \af & co & n/a & n/a & 2 hours & 26.90 & 31.85 & \textbf{25.26} & 25.84 \\
  \af & co & n/a & n/a & 4 hours & 24.88 & 25.36 & \textbf{23.66} & 30.22 \\
  \bottomrule
\end{tabular}

\end{table}

\subsection{Results and Discussion}
\label{sec:results-discussion}

In \Cref{tab:timings} we report the total PAR10 runtime across all 50 test sets
for the predicted encoding configurations from each of the six classifier
setups, four feature sets and two splitting methods.  The predicted runtimes
include the time taken to extract the features.\footnote{For features extracted
  directly from \sr (\textit{f2fsr, lipb, combi}), the feature extraction time
  added a median of 9\% (mean 21\%) to the overall running time.  The features
  extracted via \texttt{fzn2feat} added 66\% (median), 72\% (mean).}  For ease
of comparison, we report the runtime as a multiple of the virtual best time.
For example, a figure of 2.00 in \Cref{tab:timings} means that the predictions
across the 50 test sets led to a total runtime which was twice as long as the
runtime achieved if we always chose the best available configuration.  We
remind the reader that the two splitting strategies (by class vs. by instance)
yield different test sets for the 50 seeds, as explained in~\Cref{sec:splits}.

\begin{figure}[thp]
  \centering
    \includegraphics[width=0.8\textwidth]{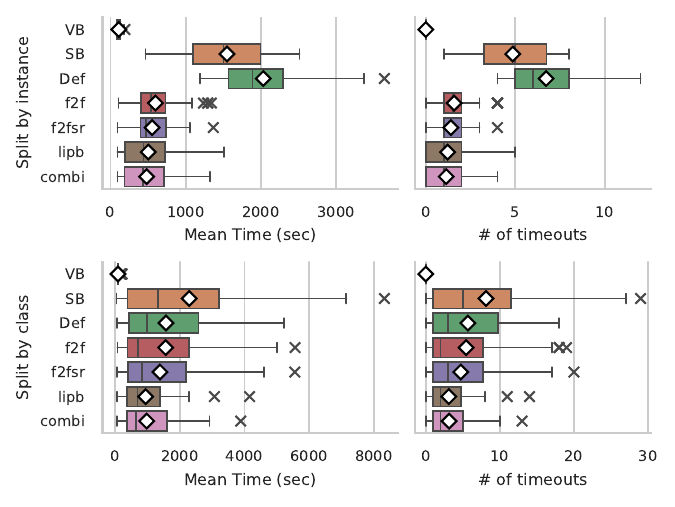}
    \caption{Prediction performance using different featuresets against
      reference times.  We show mean runtime (left) and number of timeouts
      (right) per test set across the 50 cycles, when using our preferred setup
      (\emph{pairwise combined, sample weights, custom loss}). Outliers are
      indicated with crosses and represent values more that $1.5\times$~IQR
      outside the quartiles.}
    \label{fig:pred-times}
\end{figure}

\begin{figure}[thp]
  \centering
    \includegraphics[width=\textwidth]{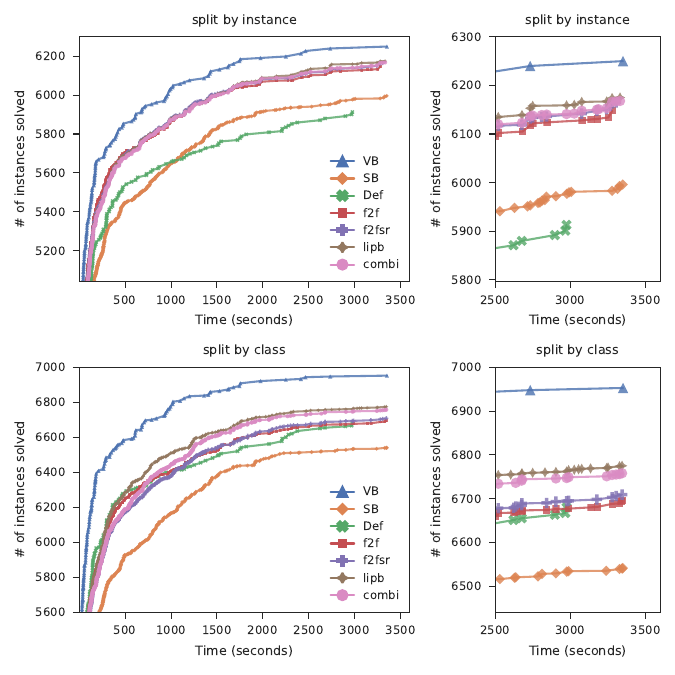}
    \caption{The number of problem instances individually solved within a given
      time for the reference selectors and our preferred predictor using
      different feature sets.  The figures on the left show the full performance
      profile; on the right we zoom in to see how many instances are solved by
      the selectors as we approach our timeout limit of 1 hour.}
    \label{fig:pred-profile}
\end{figure}

\begin{figure}[thp]
  \centering
\includegraphics[width=0.8\textwidth]{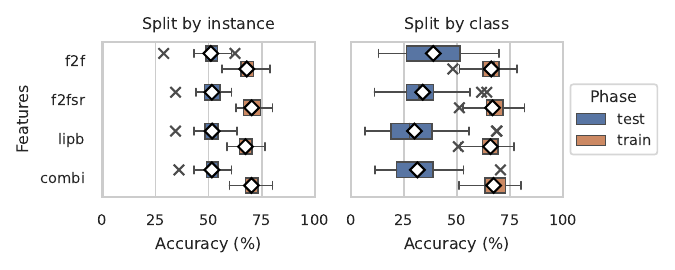}
\caption{Distributions of prediction accuracy across the 50 \emph{split, train,
  predict} cycles using our preferred setup.}
\label{fig:pred-acc}
\end{figure}

\subsubsection{On Performance and Features}
\label{sec:discuss-performance}

We found that the machine learning predictors work well, clearly outperforming
the \emph{SB} and \emph{Def} configurations.  These performance improvements can
be achieved with predictions based on the generic CSP feature sets \emph{f2f}
and \emph{f2fsr}, but are even better when using the new specialised features
(\emph{lipb}) for the majority of ML setups we implement and especially so when
predicting encodings for unseen problem classes.  Sometimes the best results are
obtained by the combined featureset \emph{combi}, again more often for unseen
problem classes.

We argue that the \emph{split-by-class} approach is both a more difficult
challenge and closer to a real-world deployment, where a new instance to solve
may belong to an unseen problem class.  However, both approaches are realistic,
so we choose the \emph{Pairwise-Voting Combined-Encoding Classifier} with
\emph{Sample Weighting} and \emph{Custom Loss} as our \emph{preferred setup} for
the rest of this paper because it is the best in the \emph{split-by-class} task
and performs competitively in the \emph{split-by-instance} setting.

In a recent survey, Kerschke et al.\ state that ``State-of-the-art per-instance
algorithm selectors for combinatorial problems have demonstrated to close
between 25\% and 96\% of the VBS-SBS gap'' \cite{kerschkeAlgoSel2019}. In these
terms, our preferred setup using \emph{lipb} features closes 59\% of the VB-SB
gap for unseen classes and 73\% for seen classes.

Because the distribution of runtimes in the split-by-class trials is skewed, we
use a non-parametric statistical test to report on the significance of the
improvement achieved by using our classifier.  We apply the Wilcoxon Signed-Rank
test for paired samples on the mean times from the SB selector choices and our
preferred selector using the \emph{lipb} features.  We obtain a $p$~value of
$6.5 \times 10^{-5}$ (well below even a $1\%$ significance level) and an effect
size of $-0.62$ using the rank-biserial correlation method which would usually
be interpreted as a medium to large effect.

\Cref{fig:pred-times} summarises the performance for our preferred setup,
showing the distribution of mean predicted PAR10 times per test set.  The mean
values are marked with diamonds and correspond to the numbers reported in
\Cref{tab:timings}, albeit not scaled.  We note that when splitting by instance,
the performance across test sets is fairly symmetrical, with very similar means
and medians for all selectors.  However, when it comes to splitting by class,
the distributions show positive skew -- this likely comes from test sets where
there are many instances from an unseen problem class for which the classifier
struggles to make the best choices.

We present an additional visualisation of selector performance
in~\Cref{fig:pred-profile}, showing the number of instances solved as time is
increased.  With both splitting strategies we observe that the featuresets
emerge in the order \emph{lipb}, \emph{combi}, \emph{f2fsr}, \emph{f2f} with the
\emph{lipb}-predicted encodings enabling the largest number of instances to be
solved within the timeout.  When splitting by instance, the four featuresets
follow a very similar trajectory; however when splitting by class a clear
advantage is shown for the specialised featuresets.  The single best (SB)
performs very differently in the two settings.  When splitting by class, there
are occasions when SB (derived from the training set) performs considerably
worse than the \emph{Tree\_Tree} default configuration, leading to the default
configuration outperforming SB.

All featuresets lead to similar performance in the split-by-instance
setting. However, when predicting for unseen problem classes, the mean runtimes
are clearly better when using the specialised featuresets.  This is also
reflected in the number of timeouts, where the \emph{lipb} featureset gives the
most robust predictions.  It is interesting that when splitting by class the
generic featureset \emph{f2f} leads to a low median runtime and low median
number of timeouts but is strongly outperformed by the specialised \emph{lipb}
features in terms of means, suggesting that the specialised features help to
avoid more costly misclassifications; i.e.\ the generic features help to ``win''
on easier problems, but don't do so well on harder ones.

A further insight is provided by \Cref{fig:pred-acc} which shows the accuracy of
predictions across the 50 training and test sets -- in this figure we see how
often the pairwise classifier ends up making exactly the ``right'' decision.  In
the split-by-instance scenario the prediction accuracy is fairly consistent
across feature sets; however, for unseen classes we observe something unexpected.
The \emph{f2f} features lead to the most accurate predictions, but, as discussed
above, the overall performance in terms of the resulting runtimes is
considerably worse.  This means that the specialised features enable the
pairwise classifier to produce a safer prediction, so that even when the
prediction made is not the absolute best encoding choice, the selected encoding
provides performance closer to the best on average.

\subsubsection{On Encoding Choices}
\label{sec:discuss-enc-choices}

\begin{figure}[thp]
  \centering
  \includegraphics[width=\textwidth]{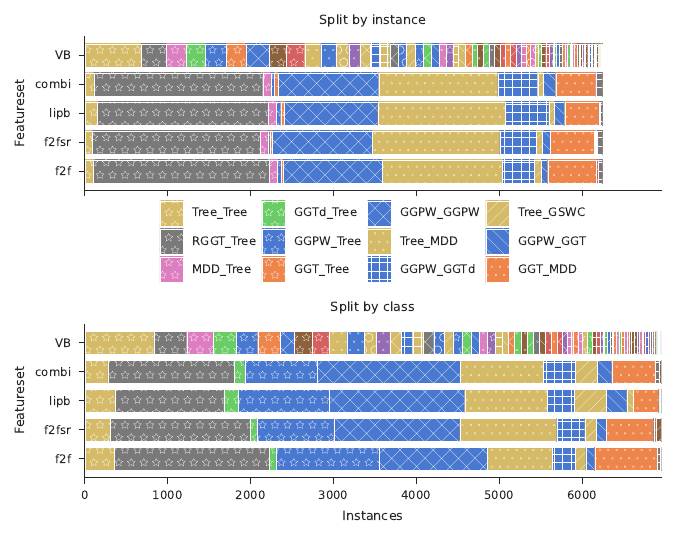}
  \caption{Frequency of each configuration (\emph{li\_pb}) selected across the 50
    test sets when using each feature set with our preferred setup. We also show
    the virtual best (VB) configuration distribution for comparison.  The colour
    indicates the LI encoding and the fill pattern shows the PB encoding.  Only
    the top 12 most used configs (of 81 in total) are shown in the legend.}
  \label{fig:pred-stacks}
\end{figure}

In \Cref{fig:pred-stacks} we show the frequency with which different encoding
configurations are predicted.  Recall that although we use a portfolio of 6
encodings, the portfolio is generated from the training set; consequently the
portfolios are different across the 50 sets.  The VB column shows a smooth
distribution of ideal encoding choices from the full range of encodings
available.  In both splitting scenarios we observe five configurations preferred
by the classifiers: \emph{RGGT\_Tree}, \emph{GGPW\_GGPW}, \emph{Tree\_MDD},
\emph{GGPW\_GGTd} and \emph{GGT\_MDD}.  Additionally, in the
split-by-class task \emph{GGPW\_Tree} is also used very often.  The GGPW
encoding for LI constraints features heavily in these choices and can therefore
be considered a very good single choice in many settings; remember that when
illustrating the building of a portfolio using the whole corpus, the
single-choice winning configuration in~\Cref{fig:portfolio-times} was
\emph{GGPW\_Tree}.  In the distribution of predictions made, there is more
variety in the PB encoding selected, with five different choices featuring in
the six top configurations mentioned.

Predictions can only be made from the portfolio determined at training time, so
in each of the 50 cycles the classifier has to choose between 6 encoding
configurations, and therefore it follows that not all configurations feature in
the predictions; in contrast the virtual best can include even edge cases where
a configuration wins only once.  Of all the encoding distributions in the
split-by-class task, it appears that the predictions made using the \emph{lipb}
features are closest to the VB in the sense of having the most even spread
amongst the configuration predictions it produces.  This is not immediately
clear from \Cref{fig:pred-stacks}.  However, when a $\chi^2$ test is carried out
with the frequencies of configuration predictions from each featureset against
the VB, the test statistics point to \emph{lipb} having the most similar spread
of choices -- the divergence measures are 27017 for \emph{lipb}, 29827 for
\emph{f2fsr}, 30277 for \emph{f2f}, and 30321 for \emph{combi}.  We suggest that
using the specialised features allows our machine learning setup to remain safe
(i.e.\ not choose encodings which lead to very bad performance) while taking
enough ``risks'' to leverage the complementarity of the portfolios by spreading
out the prediction choices more.

\subsection{Comparison with \af}
\label{sec:af}

To further assess the value of our approach, we compare with
\af~\cite{lindauerAutoFolio2015}, a sophisticated algorithm selection approach
which automatically configures algorithm selectors and ``can be applied
out-of-the-box to previously unseen algorithm selection scenarios.''  We use the
latest version of \af (the 2020-03-12 commit which adds a CSV API to the 2.1.2
release) with its default settings.  We use the algorithm selection component of
\af to make a single prediction per instance, turning on the hyperparameter
tuning option; we do not use its pre-solving schedule generation.

To compare as fairly as possible, we train \af on exactly the same training
data, and test on the same test sets as our method.  Our system takes less than
5 minutes to train using 8 cores on the cluster, so we allow \af 1 hour on one
core to tune and train.  We also run it with a more generous budget of 2 hours
and 4 hours to see if its performance improves.  The runtime performance based
on {\af}'s predictions is included in our main table of results
(\Cref{tab:timings}), in the final three rows of each main section.

Our system's predictions lead to better runtimes than \af's.  \af is designed to
be a good general algorithm selection and configuration system able to make good
predictions when choosing between different solvers.  It is likely that \af's
sophisticated decision-making is better suited to problems that run much longer
or to algorithms for which the likelihood of timeouts or non-termination is more
of an issue.  It is interesting to note that \af performs better with the
\emph{lipb} features than the generic instance features.  Allowing \af more time
for tuning leads to marginal improvement with some feature sets, but in some
cases actually results in worse performance, for example with
\emph{split-by-instance} and the \emph{combi} features.

\subsection{Feature Importance}
\label{sec:feat-imp}

\begin{figure}[tbp]
\centering
\includegraphics[width=\textwidth]{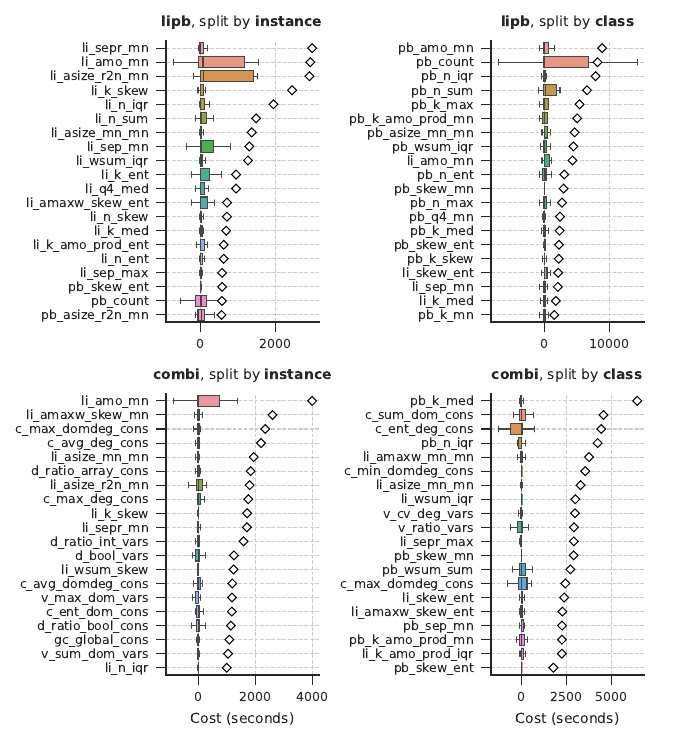}
\caption{Permutation feature importance: increase in PAR10 time over 50
  \emph{split, train, predict} cycles when each feature's data is permuted. We
  show the top 20 features according to mean importance.  We omit outliers,
  which are defined as being beyond 1.5~$\times$~IQR away from the box. The mean
  importance over the 50 cycles is shown by a diamond.  Features beginning
  \texttt{li\_} or \texttt{pb\_} refer to our LI/PB features as listed in
  \Cref{tab:features}; the other feature names refer to the generic instance
  features from the \emph{combi} feature set.}
    \label{fig:fimps}
\end{figure}

We investigate the relative importance of instance features by computing the
permutation feature importance.  Breiman~\cite{breimanRandomForests2001}
calculates ``variable importance'' in random forests by recording the percentage
increase in misclassification when each variable (feature) has its values
randomly permuted compared to when all features are used. Permuting the values
means that the distribution is preserved but the feature effectively becomes
noise.  This method is applied at prediction time to the test set, unlike the
Gini (entropy) feature importance measure which is calculated during training.
We implement this analysis but record the mean increase in PAR10 time when each
feature is permuted, effectively giving us the extra runtime cost when the
feature is lost. Each feature is randomly permuted 5 times and the mean time
increase recorded.  The distribution of feature importance thus calculated is
shown in~\Cref{fig:fimps}.  We report on the \emph{lipb} features and on the
\emph{combi} feature set which additionally contains the generic features from
\emph{f2fsr}.  We only show the top 20 features ordered by mean importance.

\begin{figure}[tbp]
\centering
\includegraphics[width=\textwidth]{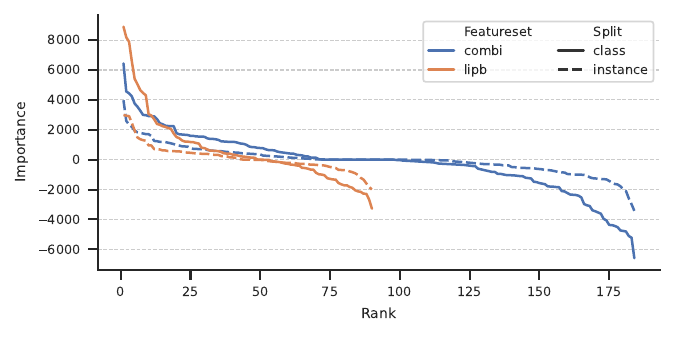}
\caption{The mean permutation feature importance across the 50 cycles for every
  feature in the \emph{lipb} and \emph{combi} featuresets, from most to least
  important.}
    \label{fig:fimps-all-means}
\end{figure}

We can see in~\Cref{fig:fimps} that for both feature sets the median feature
importance in the majority of cases is close to zero, but the mean importance
varies considerably.  This suggests that there are no features which are
dominant on their own -- most of the time a missing feature incurs no loss of
prediction performance. Indeed sometimes removing a feature can improve
performance, as shown by some negative costs in most box plots.  However, the
means of the distributions show that there are cases where each of the top
features shown is able to prevent a costly wrong choice.  The full extent of the
variation of the mean feature importance is shown
in~\Cref{fig:fimps-all-means}.

In the top 20 \emph{combi} features we find a roughly equal mix of generic
features and features specific to PB/LI constraints (the names of these features
have prefixes \texttt{pb\_} and \texttt{li\_}), when it comes to predicting for
known problem classes (i.e.\ split by instance).  This is in keeping with the
similar performance of the \emph{f2fsr} and \emph{libp} featuresets as shown
previously in \Cref{tab:timings}.  We suspect that when splitting by instance
the system is, to a large extent, recognising problem classes rather than
picking out traits of PB/LI constraints.

When we predict for unseen problem classes, the proportion of PB/LI to generic
features in the top 20 rises to 14:6, supporting the hypothesis that making
choices about which encodings to choose for certain constraint types is better
served by using features relating to those constraints in the problem instance.

Let us consider the importance of features related to PBs as distinct to LIs.
Considering the \emph{split-by-instance} case in~\Cref{fig:fimps} we note that
most of the top features relate to LIs (17 to 3); this is almost entirely
reversed when splitting by class (16 to 4 in favour of PB features).  If the
classifiers are just recognising problem classes in the \emph{split-by-instance}
case (hence predicting an encoding which has worked well for other instances in
that class), then the LI features are just as suitable as the PB features.  In
fact we saw that the generic features performed competitively in this setting.
However, when it comes to predicting for unseen problem classes, we know that
the LI/PB related features are more discriminating as shown in the performance
results (\Cref{tab:timings}).  We have also seen that there is more variation in
the best PB encoding than the best LI encoding (with GGPW often winning for LI) and
therefore the PB-related features are more prominent in this harder setting.

\begin{table}[tbp]
  \caption{The 20 most important features in our \emph{lipb} feature set by
    their mean and median permutation feature importance (PFI).  Features which
    appear in both top-20 lists are highlighted in \textbf{bold}.  These PFI
    values were obtained in the \emph{split-by-class} task and are averaged over
    the 50 \emph{split, train, predict} cycles.}
\label{tab:pfi-top}
\centering
\setlength{\tabcolsep}{3pt}
\footnotesize
\renewcommand{\arraystretch}{1}
\begin{tabular}{@{}lrrclrr@{}}
  \multicolumn{3}{c}{Top 20 by Mean} & & \multicolumn{3}{c}{Top 20 by Median}\\
  \toprule
  & \multicolumn{2}{c}{PFI (seconds)} & &  & \multicolumn{2}{c}{PFI (seconds)}\\
  \cmidrule{2-3}\cmidrule{6-7}
  Feature & Mean & Median & & Feature & Mean & Median\\
  \midrule
  \textbf{pb\_amogs\_mn} & 8899.11 & 2.15 & & \textbf{pb\_n\_sum} & 6573.40 & 114.78\\
  \textbf{pb\_count} & 8190.50 & 3.33 & & \textbf{pb\_k\_max} & 5412.10 & 68.36\\
  \textbf{pb\_n\_iqr} & 7889.19 & 0.69 & & \textbf{pb\_amogs\_size\_mn\_mn} & 4661.48 & 30.20\\
  \textbf{pb\_n\_sum} & 6573.40 & 114.78 & & pb\_k\_amogs\_prod\_ent & 1435.01 & 19.70\\
  \textbf{pb\_k\_max} & 5412.10 & 68.36 & & pb\_wsum\_sum & 591.30 & 11.05\\
  pb\_k\_amogs\_prod\_mn & 5048.18 & -0.36 & & \textbf{li\_skew\_ent} & 2118.24 & 4.61\\
  \textbf{pb\_amogs\_size\_mn\_mn} & 4661.48 & 30.20 & & pb\_n\_min & 323.83 & 3.66\\
  pb\_wsum\_iqr & 4462.16 & -0.29 & & li\_sepr\_max & -1823.14 & 3.43\\
  \textbf{li\_amogs\_mn} & 4313.62 & 1.70 & & \textbf{pb\_count} & 8190.50 & 3.33\\
  \textbf{pb\_n\_ent} & 3048.54 & 3.01 & & \textbf{pb\_n\_ent} & 3048.54 & 3.01\\
  pb\_skew\_mn & 2950.82 & 0.00 & & li\_n\_med & -164.57 & 2.22\\
  pb\_n\_max & 2703.56 & -0.21 & & \textbf{pb\_amogs\_mn} & 8899.11 & 2.15\\
  pb\_q4\_mn & 2390.52 & -0.05 & & li\_amogs\_size\_mn\_mn & 1168.76 & 2.11\\
  pb\_k\_med & 2337.28 & -0.27 & & pb\_k\_amogs\_prod\_iqr & 226.67 & 1.76\\
  \textbf{pb\_skew\_ent} & 2237.81 & 0.84 & & \textbf{li\_amogs\_mn} & 4313.62 & 1.70\\
  pb\_k\_skew & 2192.99 & -1.11 & & li\_skew\_mn & 1213.57 & 1.13\\
  \textbf{li\_skew\_ent} & 2118.24 & 4.61 & & pb\_k\_ent & 516.36 & 1.07\\
  li\_sep\_mn & 2055.11 & 0.05 & & li\_count & 422.29 & 0.87\\
  li\_k\_med & 1748.99 & 0.26 & & \textbf{pb\_skew\_ent} & 2237.81 & 0.84\\
  pb\_k\_mn & 1513.33 & 0.04 & & \textbf{pb\_n\_iqr} & 7889.19 & 0.69\\
\bottomrule
\end{tabular}
\end{table}

As a final discussion point relating to feature importance, we look in more
detail at the top 20 features from the \emph{lipb} featureset when used on
unseen problem classes.  We have commented already on the disparity between the
mean and median of the permutation feature importance.  In~\Cref{tab:pfi-top} we
list the top 20 features by mean and by median.  A positive median value tells
us that a feature is more often than not valuable in making good predictions.
The mean indicates the overall contribution in a different way, i.e.\ how much
time is lost on average per prediction batch across the 50 cycles.  The features
highlighted in bold type appear in both top-20 lists and so can help to explain
what kinds of features are most helpful.  We note that these mostly pertain to
size, e.g. \texttt{pb\_n\_sum} is the total number of terms across all PBs,
\texttt{pb\_k\_max} is the highest upper bound $k$,
\texttt{pb\_amogs\_size\_mn\_mn} is the mean of the mean size of the AMO groups
in PBs, \texttt{pb\_count} is the number of PB constraints.

There are limitations to how much we can read into the permutation feature
importance (PFI), in particular because of two factors: PFI considers features
in isolation and we have quite a large number of features. A feature $\alpha$
may be discriminating but could be masked by another feature $\beta$ with which
it is highly correlated, so when we permute {$\alpha$}'s values, there may not
be a great loss in prediction performance while the information from $\beta$
remains.  Thus we could wrongly conclude that $\alpha$ is not very valuable.  We
have also shown that the features in \emph{libp} and \emph{f2fsr} can give
comparable prediction performance (especially when splitting by instance) even
though they consider different aspects of a CSP.

It is very difficult to draw strong conclusions about which features are the
most significant.  Many algorithms exist to aid feature selection before
applying machine learning methods.  Although further work in reducing
the featuresets could be of value, we have shown that better predictions are
achievable when using only the constraint-specific features in the split-by-class setting.

\subsection{Analysis of the Configuration Space}
\label{sec:conf-space}

We end our account of the empirical investigation with a brief analysis of
the configuration space in which we are making encoding choices.  We have argued
already that the task of selecting suitable SAT encodings is not just a simple
classification task.  We hope the observations in this section shed some light
on important considerations when selecting encodings for a set of problems.

\begin{table}[p!]
  \caption{Summary of the 20 best encoding configurations across the problem
    corpus by two different criteria. \emph{Left}: the encodings which are best
    most frequently, showing the number of ``wins'' and the mean runtime for the
    instances on which it wins; the mean is rounded to the nearest second.
    \emph{Right}: the encodings whose allocated instances in the virtual best
    selection have the highest total runtimes, and their contribution as a
    percentage of the total VB runtime.  Encodings appearing in both top 20
    lists are highlighted in \textbf{bold} type.}
  \label{tab:conf-space}
  \centering
  \begin{tabular}{@{}lrrllr@{}}
    \toprule
    \multicolumn{3}{c}{Most Frequent Winners}&&\multicolumn{2}{c}{Biggest Contributions to VB}\\
    \cmidrule(r){1-3}\cmidrule(l){5-6}
    Encodings (LI\_PB) & Wins & Mean Time & & Encodings (LI\_PB) & \% of VB\\
    \cmidrule(r){1-3}\cmidrule(l){5-6}

    \textbf{Tree\_Tree} & 73 &  46 & & \textbf{GGPW\_GGPW} & 23.2 \\
    \textbf{RGGT\_Tree} & 31 & 127 & & \textbf{RGGT\_Tree} & 5.7 \\
    \textbf{GGPW\_GGPW} & 26 & 611 & & \textbf{GPW\_Tree} & 5.4 \\
    MDD\_Tree & 26 &  30 & & \textbf{GGPW\_GGT} & 5.3 \\
    \textbf{GPW\_Tree} & 25 & 148 & & GGT\_MDD & 5.0 \\
    GGTd\_Tree & 24 &  28 & & \textbf{Tree\_Tree} & 5.0 \\
    GLPW\_Tree & 21 &   7 & & \textbf{Tree\_MDD} & 4.3 \\
    GGT\_Tree & 21 &  29 & & \textbf{GGPW\_RGGT} & 3.3 \\
    GSWC\_Tree & 19 &  17 & & GSWC\_GGT & 2.8 \\
    \textbf{Tree\_MDD} & 19 & 153 & & GGPW\_GMTO & 2.7 \\
    GGPW\_MDD & 18 &  16 & & GLPW\_MDD & 2.6 \\
    Tree\_RGGT & 15 &  21 & & GSWC\_GSWC & 2.6 \\
    GMTO\_MDD & 14 &  57 & & GLPW\_GGTd & 2.3 \\
    \textbf{Tree\_GPW} & 13 &  77 & & GGT\_RGGT & 1.6 \\
    Tree\_GGTd & 11 &  76 & & GSWC\_GGPW & 1.6 \\ 
    GGPW\_GGTd & 11 &  76 & & MDD\_GGTd & 1.6 \\
    Tree\_GSWC & 11 &  43 & & GGT\_GGPW & 1.5 \\
    \textbf{GGPW\_RGGT} & 10 & 223 & & GGTd\_GSWC & 1.5 \\
    \textbf{GGPW\_GGT} & 10 & 362 & & \textbf{Tree\_GPW} & 1.5 \\
    RGGT\_GGPW & 10 &   7 & & MDD\_GMTO & 1.4 \\
    \bottomrule
  \end{tabular}

\end{table}

\begin{figure}[p!]
\includegraphics[width=0.9\textwidth]{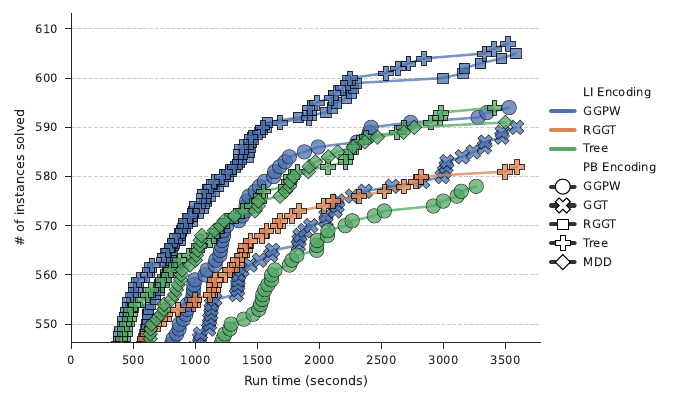}
\caption{Performance profile of selected encodings on the entire corpus, showing
  how many problems can be individually solved within a given time, up to the
  timeout of 1 hour.  The LI encoding is represented by the line colours and the
  PB encoding by the marker shape. We show the encoding configurations which
  appear in the top 20 both in terms of their contribution to VB and the number
  of times they are the best (see~\Cref{tab:conf-space}).}
\label{fig:conf-space}
\end{figure}

In \Cref{tab:conf-space} we list the 20 best performing encoding configurations
across the entire cleaned corpus\footnote{Here we are considering each problem
  instance in the corpus once, not sampling repeatedly.  Recall also that the
  cleaned corpus only contains instances for which at least one encoding
  configuration terminates before the timeout, so the PAR10 penalty does not
  apply here.} using two different criteria.  Firstly, according to how often an
encoding configuration is the best available; secondly, calculating the
proportion of the total VB runtime allocated to a configuration.  For instance,
we see that \emph{Tree\_Tree} is the clear winner in the former league table,
with more than twice as many wins as the next entry (73 vs. 31).  However, the
instances on which it wins have a mean runtime of 46 seconds.  We can calculate
its contribution to the VB as roughly $73 \times 46 \approx 3400 s$,
approximate because the mean is rounded.  On the other hand, \emph{RGGT\_Tree}
wins fewer times but the relevant instances are almost three times harder with a
mean runtime of 127 seconds, making a contribution to the VB of approximately
$31 \times 127 \approx 3900 s$.

\Cref{fig:conf-space} shows the performance profile of the encoding
configurations which appear in both top 20 lists, and highlighted in bold in
\Cref{tab:conf-space}.  As before, we see that GGPW is a great choice for LI
constraints, appearing in 4 of the top 8 combined performers, whereas there is
greater variety in the PB encodings in the best configurations.
\Cref{fig:conf-space} also demonstrates that these ``top'' encoding choices are
excellent all-rounders -- the top 2 are each able to solve over 600 of the 614
instances in the cleaned corpus (in which every instance is solvable within the
timeout by at least one encoding configuration from the full set of 81).

In terms of prediction accuracy, choosing \emph{Tree\_Tree} most often makes
sense, and indeed this is the default encoding provided by \sr.  This is an
excellent choice if the task is to solve very many problems, each of which are
individually relatively easy, i.e.\ would solve in under a minute on our
hardware.  If, instead, the task is to solve ``harder'' problems, then, at least
according to our corpus of problems, GGPW is a good choice for both LI and PB
constraints.  Recall that all encodings except \emph{Tree} take advantage of AMO
groups to reduce the size of the SAT encoding -- another important consideration
when selecting an encoding.


\section{Related Work}

In recent work, new or improved SAT encodings of linear
constraints~\cite{abioEncLinCons2020} and pseudo-Boolean constraints (combined
with AMO constraints)~\cite{bofill2019sat,bofillPBAMO2022} have been devised and
their performance compared on several benchmark problems. The scaling properties
of encodings are studied, and it is suggested that smaller encodings should be
used when coefficients or values of integer variables are large.  However, to
the best of our knowledge the problem of selecting an encoding (particularly for
a previously-unseen problem class) has not been systematically addressed for LI
or PB constraints.  We use the full set of encodings from one recent
paper~\cite{bofillPBAMO2022} combined with automatic AMO
detection~\cite{ansocp2019autodetect}.

MeSAT~\cite{stojadinovicMeSAT2014} and Proteus~\cite{hurleyProteus2014} both
select SAT encodings using machine learning.  MeSAT has two encodings of LI
constraints: the order encoding~\cite{tamura2009compiling}; and an encoding
based on enumeration of allowed tuples of values (which uses a direct encoding
of the CSP variables). It is not clear whether high-arity sums are broken up
before encoding. MeSAT selects from three configurations using a k-nearest
neighbour classifier using 70 CSP instance features. They report high accuracy
(within 4\% of the virtual best configuration), however the single best
configuration is only 18\% slower than the virtual best.  Proteus makes a
sequence of decisions: whether to use CSP or SAT; the SAT encoding; and the SAT
solver to use. The portfolio contains three SAT encodings: direct, support, and
a hybrid direct-order, however the encoding of LI constraints is not
specified~\cite{hurleyProteus2014}. Proteus generates each candidate SAT
encoding and extracts features of the SAT formula to inform its selection --
scaling this approach would be difficult when several constraint types are
involved, each with many encoding choices. Results show that the choice of
encoding (combined with the choice of SAT solver) is important and that machine
learning methods can be effective in their context.

Soh et al~\cite{soh2015hybrid} have proposed a hybrid encoding of CSP to SAT in
which each variable may have the log or order encoding (but not both). A
hand-crafted heuristic is used to automatically select one of the two encodings
for each variable.  A new encoding is defined for linear inequalities that
contain a mix of log and order encoded integer variables.  The hybrid encoding
is shown to outperform both log and order encodings, demonstrating the potential
of selecting encodings for individual variables or constraints rather than for
an instance.  Log encodings have been shown to give good performance, with the
Picat-SAT solver~\cite{zhou2017optimizing} placing highly in recent CP challenges -- Picat-SAT uses a log
encoding for integer variables. Our set of encodings includes two that use
log arithmetic internally (GGPW and GMTO), however none 
employ a log encoding of integer variables. Appendix~\ref{sec:log-encs} contains
a brief comparison to Picat-SAT using selected instances with large domains 
from our benchmark set.


\section{Conclusions and Future Work}

We have shown that it is possible to close much of the performance gap between
the single best and virtual best SAT encodings by using machine learning to
select encoding configurations based on instance features.  We have studied the
problem of selecting encodings for instances of previously-unseen classes, a
problem that is more challenging and arguably more realistic than the usual
setting where training and test instances are drawn from the same set of problem
classes.  General instance features such as those provided by
\texttt{fzn2feat}~\cite{amadiniFeatEx2014} perform well; however the
introduction of features specific to linear integer and pseudo-Boolean
constraints has enabled us to improve the quality of predictions.

We describe a machine learning method that performs well, and investigate
several variations of it.  We presented a thorough experimental analysis of the
method. Our comparison with \af shows that our method is much more effective
on the specific task in hand than the competition-winning algorithm selector
with its more generalised capabilities.

We calculate feature importance values and discuss the relative importance of
features from different featuresets as well as within the specialised
\emph{lipb} featureset.  We find that in these specialised features, the
features of PB encodings made more difference than those of LI encodings, partly
because the GGPW encoding was a frequent best choice for LI, whereas the best PB
encoding varied more.

We intend to build on these results by considering other constraint types for
which multiple SAT encodings exist.  It may also be beneficial to expand the
problem corpus to have a more even distribution of problem instances per class
and to broaden the range of constraint models represented.

\section*{Declarations}

\subsubsection*{Acknowledgments} We are very grateful to Nguyen Dang for helpful
conversations about portfolio approaches.  This project was undertaken on the
Viking Cluster, a high performance compute facility provided by the University
of York. We are grateful for computational support from the University of York
High Performance Computing service, Viking and the Research Computing team.

\subsection*{Funding}
This research was supported by EPSRC grants EP/W001977/1 
and EP/R513386/1. 

\subsection*{Availability of data and materials}
The constraint models and experimental results are available at
\url{https://github.com/felixvuo/lease-data}.

\subsection*{Code availability }
The source code for running the experiments and evaluating the results is also
available at \url{https://github.com/felixvuo/lease-data}.

\subsection*{Authors' contributions}

With reference to the Contributor Roles Taxonomy (CRediT):
\begin{itemize}
\item Felix Ulrich-Oltean: conceptualization, data curation, formal analysis,
  investigation, methodology, project administration, software, visualization,
  writing - original draft, review \& editing
\item Peter Nightingale: conceptualization, formal analysis, funding
  acquisition, methodology, project administration, software, supervision,
  writing - original draft, review \& editing
\item James Alfred Walker: formal analysis, methodology, supervision, writing -
  review \& editing
\end{itemize}

\bibliographystyle{unsrturl}
\bibliography{references}

\appendix

\section{On Log Encodings}
\label{sec:log-encs}

Logarithmic encodings have been shown to be competitive for encoding linear
constraints to SAT, in particular faring well when large domains are involved
\cite{zhou2017optimizing}. Two of the encodings considered here (GGPW and GMTO)
use some form of log encoding internally, however the domains of decision
variables are not log encoded.

We carried out a small informal experiment to investigate the potential benefits
of using a log encoding for both variables and constraints, as implemented in
Picat-SAT \cite{zhou2017optimizing} in Picat 3.5.  We chose two problem classes
with potentially large integer domains and solved them using Picat-SAT, both by
writing the model in the Picat language and by exporting a FlatZinc file from
Savile Row.  The experiment was run on a PC laptop with a i5-1135G7 2.40GHz
processor and 16GB RAM.  The results are shown in \Cref{tab:log-enc}. Details of
models and instances can be found in the experimental repository.

The Grocery problem contains large integer domains with both sum and product
constraints, and the log encoding clearly performs much better in this case. For
Knapsack, the GGPW and Tree encodings are competitive and perform better except
on the 30-large instance where all item weights and values have been
artificially scaled up by a factor of 10 for this experiment. In summary, log
encoding of domains would seem to be a valuable additional encoding choice to
consider in future work.

\begin{table}[htb]
  \caption{Total runtime in seconds for six problem instances using different
    solvers and SAT encodings: \emph{Picat} uses the model description in the
    Picat language with the \texttt{sat} module, \emph{Fzn-Pi} uses Savile Row
    to produce a FlatZinc file which is solved using the
    \texttt{fzn\_picat\_sat} program, \emph{GGPW-GGPW} uses Savile Row with
    Kissat and encodes both linear and PB constraints using the GGPW encoding,
    \emph{GGPW-Tree} uses Savile Row with Kissat and encodes linear constraints
    with GGPW and PB constraints using the default Tree encoding.  Note that in
    the grocery problem there are no PB constraints, so the choice of PB
    encoding is irrelevant and we show only \emph{GGPW-GGPW}.
    $\lvert{D}\rvert_{\max}$ indicates the largest domain size in each instance.}

  \label{tab:log-enc}
  \centering
  \begin{tabular}{@{}lrrrrr@{}}
    \toprule
    Problem & $\lvert{D}\rvert_{\max}$ & \multicolumn{4}{c}{Total runtime (seconds)} \\
    \midrule
            & & Picat & Fzn-Pi & GGPW-GGPW & GGPW-Tree \\
    \cmidrule{3-6}
    grocery, target-644 & 1380001 & 0.42 & \textbf{0.04} &  3.87 & --- \\
    grocery, target-675 & 6270665 & \textbf{0.09} & 0.24 & 30.99 & --- \\
    grocery, target-713 & 5562469 & 0.30 & \textbf{0.05} & 18.30 & --- \\
    knapsack, 30-large & 1189 & \textbf{16.17} & 19.30 & 73.33 & 49.38 \\
    knapsack, 59-items & 1457 & 13.99 & 14.37 &  \textbf{7.82} &  9.31 \\
    knapsack, 81-items & 2497 & 84.75 & 93.69 & 72.75 & \textbf{52.31} \\
    \bottomrule
  \end{tabular}
\end{table}

\end{document}

%% file: diagram.tex
  \tikzset{%
    inout/.style={draw=ourcolsGray,rectangle,fill=white,
      minimum width=2em,minimum height=15pt,text width=10em,text centered},
    stafin/.style={draw=ourcolsGray,rectangle,fill=white,
      rounded corners,thick,
      minimum width=2em,minimum height=15pt,text width=10em,text centered},
    proc/.style={draw=none,rectangle,fill=ourcolsLightGray, minimum width=2em, minimum
      height=15pt,text width=10em,text centered},
    block/.style={draw=ourcolsLineGray,dashed,thick,fill=none,inner sep=1ex},
    arrow/.style = {thick,->,>=stealth},
    line/.style = {thick,--,>=stealth}
  }
  \begin{tikzpicture}[node distance=2.4ex]
    
    \node (inst) [stafin,text width=15em] {problem
      instances\\ (\texttt{.eprime} models, \texttt{.param} files)};
    
    \node (srext) [proc,below=of inst] {\sr extract features};

    \node (srsolv) [proc,left=of srext] {\sr solve with SAT backend};

    \node (srfzn) [proc,right=of srext] {\sr output flatzinc};

    \node (fzn2) [proc,below=of srfzn,text width=10ex] {\texttt{fzn2feat}};

    \node (fetim) [inout,below=of fzn2] {feature extraction times};

    \node (feats) [inout,left=of fetim] {features};

    \node (soltim) [inout,left=of feats] {solving times};

    \node (clean) [proc,below=of soltim,xshift=6em,text width=15em] {drop instances without
      PB/LI or where all encodings time out;\\apply PAR10};

    \node (dset) [inout,below=of clean] {cleaned dataset};

    \node (calref) [proc] at (dset-|fetim) {calculate reference and
      prediction times};

    \node (split) [proc,below=of dset] {train/test split};

    \node (trnset) [inout,below=of split,xshift=-6em] {training set};

    \node (tstset) [inout,right=of trnset] {test set};

    \node (selpf) [proc,below=of trnset] {select best 6 encoding configurations};
    
    \node (pfolio) [inout,below=of selpf] {portfolio};

    \node (trainc) [proc,below=of pfolio] {train classifiers};

    \node (clfs) [inout,below=of trainc] {trained classifiers};

    \node (pred) [proc,right=of clfs] {predict and vote};

    \node (preds) [inout] at (pfolio-|calref) {predicted encoding configurations};

    \node (agg) [stafin,right=of pred,yshift=-2em] {aggregated prediction data
      for analysis};
    
    \node (cycles) [block,fit=(split) (clfs) (pred)] {};

    \node [below,text width=18em,text centered] at (cycles.south) {\textit{for
    each splitting method and setup,\\perform 50 cycles with different seeds}};

    \draw [arrow] (inst) -- (srsolv);
    \draw [arrow] (inst) -- (srfzn);
    \draw [arrow] (inst) -- (srext);
    \draw [arrow] (srfzn) -- (fzn2);
    \draw [arrow] (fzn2) -- (feats);
    \draw [arrow] (fzn2) -- (fetim);
    \draw [arrow] (srsolv) -- (soltim);
    \draw [arrow] (srext) -- (feats);
    \draw [arrow] (srext) -- (fetim);
    \draw [arrow] (feats) -- (clean);
    \draw [arrow] (soltim) -- (clean);
    \draw [arrow] (clean) -- (dset);
    \draw [arrow] (dset) -- (split);
    \draw [arrow] (split) -- (trnset);
    \draw [arrow] (split) -- (tstset);
    \draw [arrow] (trnset) -- (selpf);
    \draw [arrow] (selpf) -- (pfolio);
    \draw [arrow] (pfolio) -- (trainc);
    \draw [arrow] (trainc) -- (clfs);
    \draw [arrow] (tstset) -- (pred);
    \draw [arrow] (clfs) -- (pred);
    \draw [arrow] (pred) -- (preds);
    \draw [arrow] (preds) -- (calref);
    \draw [arrow] (fetim) -- (calref);
    \draw [arrow] (dset) -- (calref);
    \draw [arrow] (calref) -- ++(2.5,0) |- (agg);
  \end{tikzpicture}